\UseRawInputEncoding
\documentclass[10pt,journal,cspaper,compsoc]{IEEEtran}

\usepackage{times}
\usepackage{epsfig}
\usepackage{graphicx}
\usepackage{amsmath}
\usepackage{amssymb}
\usepackage{bm}
\usepackage{mathrsfs}
\usepackage{txfonts}
\usepackage{algorithm} 
\usepackage{enumerate}
\usepackage{multirow}

\usepackage{epstopdf}
\usepackage{setspace}
\usepackage{epsfig}

\usepackage[left=0.5in,right=0.5in,top=0.3in,bottom=0.5in]{geometry} 

\usepackage{amssymb}
\usepackage{pifont}
\newcommand{\cmark}{\ding{51}}%

\usepackage[table]{xcolor}

\ifCLASSOPTIONcompsoc
\usepackage[nocompress]{cite}
\else
\usepackage{cite}
\fi

\ifCLASSINFOpdf
\else
\fi

\usepackage[pagebackref=true,breaklinks=true,letterpaper=true,colorlinks,bookmarks=false]{hyperref}

\usepackage{color}
\definecolor{myColor}{rgb}{0, 0, 1}

\newcommand{\etal}{{\it et al. }}

\usepackage{multirow}

\usepackage[breaklinks=true,bookmarks=false]{hyperref}

\newcommand{\ie}{\textit{i}.\textit{e}.}

\usepackage{subcaption,booktabs}
\begin{document}
	%
	\title{From NeRFLiX to NeRFLiX++: A General NeRF-Agnostic Restorer Paradigm}
	%
	%
	%
	%
	%
	
	\author{Kun~Zhou,
		Wenbo Li,
		Nianjuan~Jiang,~\IEEEmembership{Member,~IEEE,}
		Xiaoguang~Han,~\IEEEmembership{Member,~IEEE,}
		and~Jiangbo~Lu,~\IEEEmembership{Senior~Member,~IEEE}
		\IEEEcompsocitemizethanks{\IEEEcompsocthanksitem Kun Zhou, Nianjuan Jiang are with SmartMore Corporation, Ltd., Shenzhen, China. 
			E-mail:\{zhoukun303808,jnianjuan\}@gmail.com.
			\IEEEcompsocthanksitem Wenbo Li is with CSE, The Chinese University of Hong Kong, HongKong, China. E-mail: wenboli@cse.cuhk.edu.hk.
			\IEEEcompsocthanksitem Xiaoguang Han is with Shenzhen Institute of Big Data, The Chinese University of Hong Kong (Shenzhen), Shenzhen, China. E-mail: hanxiaoguang@cuhk.edu.cn.
			\IEEEcompsocthanksitem Jiangbo Lu is with SmartMore Corporation, Ltd., Shenzhen, China, and also with South China University of Technology, Guangzhou 510641, China. E-mail: jiangbo.lu@gmail.com.  (Corresponding author: Jiangbo Lu.)}
	}

	\IEEEtitleabstractindextext{
		\begin{abstract}
			Neural radiance fields~(NeRF) have shown great success in novel view synthesis. However, recovering high-quality details from real-world scenes is still challenging for the existing NeRF-based approaches, due to the potential imperfect calibration information and scene representation inaccuracy. Even with high-quality training frames, the synthetic novel views produced by NeRF models still suffer from notable rendering artifacts, such as noise and blur. To address this, we propose NeRFLiX, a general NeRF-agnostic restorer paradigm that learns a degradation-driven inter-viewpoint mixer. Specially, we design a NeRF-style degradation modeling approach and construct large-scale training data, enabling the possibility of effectively removing NeRF-native rendering artifacts for deep neural networks. Moreover, beyond the degradation removal, we propose an inter-viewpoint aggregation framework that fuses highly related high-quality training images, pushing the performance of cutting-edge NeRF models to entirely new levels and producing highly photo-realistic synthetic views. Based on this paradigm, we further present NeRFLiX++ with a stronger two-stage NeRF degradation simulator and a faster inter-viewpoint mixer, achieving superior performance with significantly improved computational efficiency. Notably, NeRFLiX++ is capable of restoring photo-realistic ultra-high-resolution outputs from noisy low-resolution NeRF-rendered views. Extensive experiments demonstrate the excellent restoration ability of NeRFLiX++ on various novel view synthesis benchmarks.
		\end{abstract}
		
		}

	\maketitle
	
	\IEEEdisplaynontitleabstractindextext

	%
	\IEEEpeerreviewmaketitle
	
	
	\begin{figure*}[t]
		\centering
		
		\includegraphics[width=1.0\linewidth]{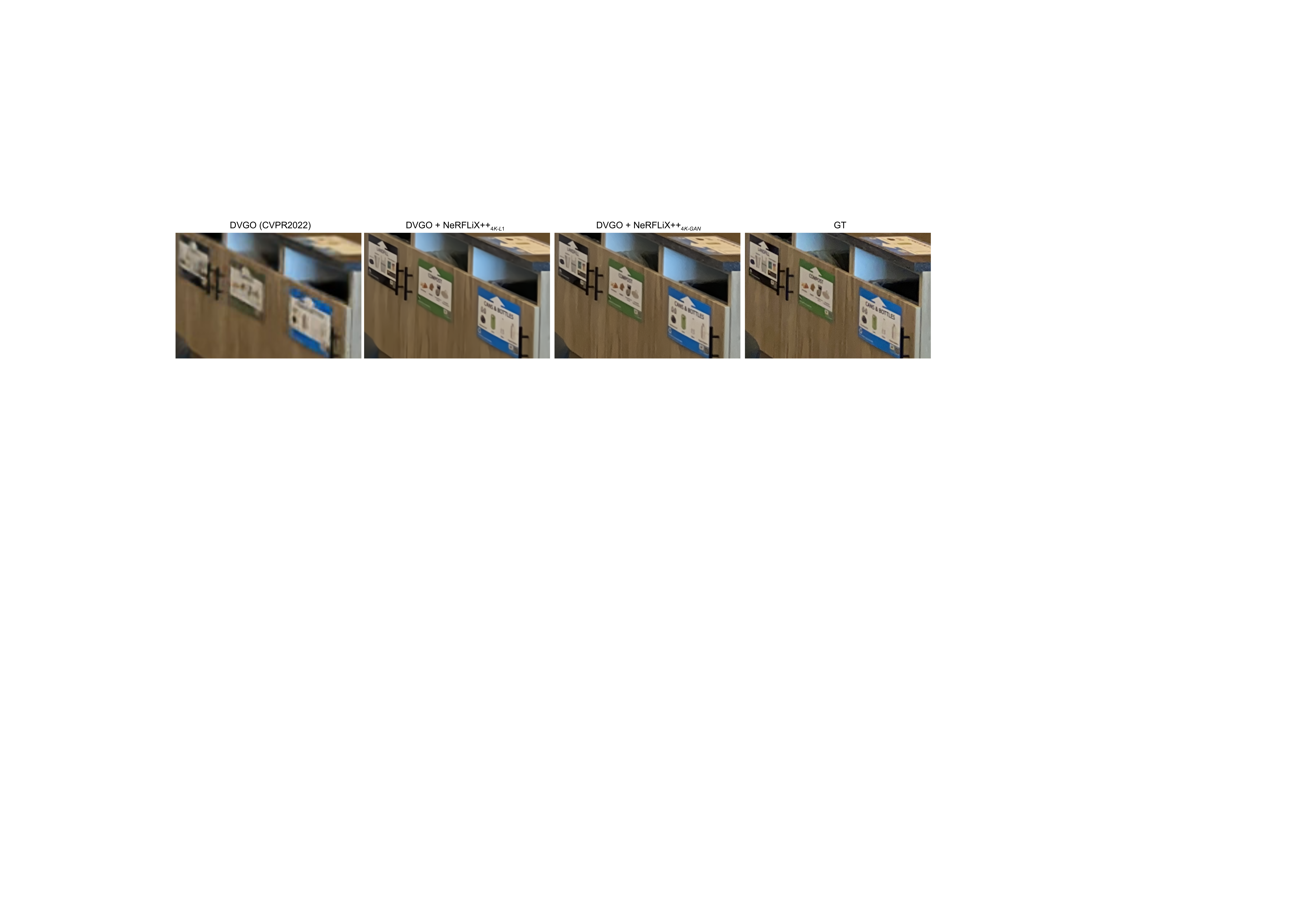} 
		\caption{Visualization of restoration results of our proposed NeRFLiX++ for 4K images. It is clear that NeRFLiX++ produces photo-realistic 4K frames from low-resolution and noisy inputs rendered by DVGO~\cite{Sun_2022_CVPR}. } 
		\label{fig:figteasing}
	\end{figure*} %
	
	\IEEEraisesectionheading{\section{Introduction}\label{sec:introduction}}
	\IEEEPARstart{P}{hoto}-realistic novel view synthesis is a long-standing problem in the fields of computer vision and graphics. Recent years have seen the emergence of learning-based approaches, such as NeRF (Neural Radiance Fields) and its follow-ups, which utilize neural networks to represent 3D scenes and employ various rendering techniques to synthesize novel views. To achieve high-quality rendering, it is essential to design physically-aware systems that optimize multiple factors, including geometry, environment lighting, object materials, and camera poses. However, despite advancements, state-of-the-art NeRF models may still suffer from undesirable rendering artifacts when relying solely on a limited number of input views, as discussed in~\cite{zhang2021nerfactor,guo2022nerfren,zhang2022modeling,lin2021barf,jeong2021self,wu2022dof}.

	Towards high-quality novel view synthesis, we propose NeRFLiX~\cite{Zhou_2023_CVPR} that delivers pioneering efforts to investigate the feasibility of simulating large-scale NeRF-style paired data for training a NeRF-agnostic restorer. The system comprises two primary components: (1) a NeRF-style degradation simulator (NDS) and (2) an inter-viewpoint mixer (IVM). Inspired by practical image restoration approaches~\cite{zhang2021designing,wang2021real}, NeRFLiX systematically analyzes typical NeRF rendering artifacts and presents three manually designed degradations to simulate NeRF-rendered noises. We take advantage of NDS to generate a substantial amount of simulated training data and further develop a deep restorer, \ie, IVM, to remove NeRF-style artifacts. Consequently, NeRFLiX demonstrates remarkable performance in synthesizing novel views of high fidelity, thereby extending the capabilities of NeRF models to new frontiers. However, there are two perspectives that deserve further investigation: (1) the inadequacy of manual degradation designs in accounting for the dispersion of real NeRF-rendered artifacts, and (2) the difficulty of employing the large inter-viewpoint mixer for processing high-resolution frames. 
	
	Hereafter, we extend NeRFLiX to NeRFLiX++ by introducing a two-stage degradation simulation approach, combined with a more efficient guided inter-viewpoint mixer. This refined framework not only achieves superior or comparable performance but also demonstrates significantly improved inference efficiency.
	
	
	%
	
	
	\vspace{0.03in}
	\noindent\textbf{Two-stage degradation simulation.} To bridge the domain gap between NeRF-rendered artifacts and simulated ones, we propose a two-stage degradation simulation scheme that consists of a hand-crafted degradation simulator and a deep generative degradation simulator.  In the first stage, we utilize a similar degradation pipeline as NeRFLiX, but incorporate more basic degradations~(\ie, illumination jetting and brightness compression) to obtain an initially degraded frame. In the second stage, we leverage generative adversarial training to optimize the output from the first stage, making it statistically closer to NeRF-rendered views. However, training a deep generative network for our approach is challenging due to limited samples in the target domain. We observe that conventional pixel-to-pixel supervision actually constrains the diversity of simulated noise. To address this issue, we draw inspiration from Beby-GAN~\cite{li2022best} and propose a novel approach that leverages image self-similarity and introduces a weighted top-$K$ buddy loss for adversarial training. Specifically, given a simulated patch, we search for the $K$ most relevant ``buddies" (image patches) from the real sample (NeRF-rendered image), which are then used to provide weak supervision. This approach significantly enhances the diversity of generated patterns, resulting in improved degradation modeling. With the two-stage simulator, we are able to construct sizable training pairs and demonstrate that various deep restorers can be trained to effectively eliminate NeRF-style artifacts.

	
	\vspace{0.03in}
	\noindent\textbf{Guided inter-viewpoint mixer.} To overcome the efficiency challenges of handling high-resolution frames for NeRFLiX, which incorporates a recurrent aggregation strategy to fuse details from reference views, we propose a more efficient guided inter-viewpoint aggregation scheme in NeRFLiX++. We achieve this by first estimating dense pixel-wise correspondences (optical flow) at a low resolution, based on several considerations. Firstly, the down-sampling operation results in smaller displacements between images, which lowers the difficulty of estimation. Secondly, the distributions of the rendered view and reference views become closer, resulting in more accurate correspondence estimation. Lastly, this approach is computationally more efficient. We then employ a coarse-to-fine guided aggregation by leveraging motion fields predicted at lower scales to aggregate information at higher scales. This strategy eliminates the need for recurrent high-resolution correspondence estimation, largely improving computational efficiency. Compared with NeRFLiX, NeRFLiX++ achieves superior results on benchmark datasets, such as Tanks and Temples and Noisy LLFF Synthetic, while performing on par with the LLFF dataset. Notably, NeRFLiX++ is 9.2$\times$ faster in processing scenes of a $1024 \times 1024$ size, highlighting its significant efficiency improvements. 
	
	In summary, our contributions are threefold:
	\begin{itemize}
		\item 
		\textbf{Accurate NeRF degradation modeling. } We propose a two-stage degradation modeling scheme that closely approximates the statistical characteristics of real NeRF-rendered artifacts. Through this scheme, we demonstrate the effectiveness of existing deep image/video restorers and our proposed NeRFLiX/NeRFLiX++ in further enhancing the quality of NeRF-rendered views using simulated samples.
		
		\item
		\textbf{Efficient inter-viewpoint mixer.} We develop an efficient inter-viewpoint aggregation method that effectively integrates information from multiple viewpoints, enabling fast and accurate processing of ultra-high-resolution frames.
		
		\item 
		\textbf{High-quality super-resolution.}
		Given the high efficiency of our accelerated inter-viewpoint aggregation, we demonstrate the potential of NeRFLiX++ to be extended to super-resolution tasks, generating photo-realistic 4K frames from noisy 1K NeRF-rendered views, as illustrated in Fig.~\ref{fig:figteasing}.
		
	\end{itemize}

	A preliminary version of our work, NeRFLiX~\cite{Zhou_2023_CVPR}, has been accepted at the IEEE/CVF Conference on Computer Vision (CVPR) 2023. This extended version presents several key contributions and advancements. First, we address the limitations of hand-crafted degradations by introducing a novel two-stage degradation scheme that better models the complex distribution of NeRF-rendered frames. Second, we systematically analyze the efficiency of the recurrent inter-viewpoint mixer and propose a faster alternative. These improvements result in NeRFLiX++ achieving superior performance with significantly reduced computational costs. Moreover, we demonstrate that NeRFLiX++ can be easily applied to super-resolving photo-realistic 4K images from low-resolution NeRF-rendered views with minimal architecture modifications. The code of NeRFLiX++ will be released at \url{https://redrock303.github.io/nerflix_plus/} to facilitate future research.

	\section{Related Work}\label{sec:relatedWork}
	In this section, we review the relevant approaches consisting of NeRF-based novel view synthesis, degradation simulation in low-level version, and inter-frame correspondence estimation.

	\vspace{0.03in}
	\noindent\textbf{NeRF-based novel view synthesis.}
	This field has received a lot of attention recently and has been thoroughly investigated. For the first time, Mildenhall~\etal ~\cite{mildenhall2020nerf} propose the neural radiance field to implicitly represent static 3D scenes and synthesize novel views from multiple posed images. Inspired by their successes, a lot of NeRF-based models~\cite{cole2021differentiable,tancik2022block,barron2021mip,yang2022recursive,pumarola2021d,martin2021nerf,wang2021ibrnet,hu2022efficientnerf,mildenhall2022nerf,xiang2021neutex,guo2022nerfren,mueller2022instant,suhail2022light,kurz2022adanerf,johari2022geonerf,ichnowski2021dex,lin2021efficient,chen2021mvsnerf,deng2022compressing,zhang2021physg,wang2022nerf,rebain2021derf,zhang2022fast} have been proposed. For example, point-NeRF~\cite{xu2022point} and DS-NeRF~\cite{deng2022depth} incorporate sparse 3D point cloud and depth information for eliminating the geometry ambiguity of NeRFs, achieving more accurate and efficient 3D point sampling as well as better rendering quality. Plenoxels~\cite{fridovich2022plenoxels}, TensoRF~\cite{tensorf}, DirectVoxGo~\cite{Sun_2022_CVPR}, FastNeRF~\cite{garbin2021fastnerf}, Plenoctrees~\cite{yu2021plenoctrees}, KiloNeRF~\cite{reiser2021kilonerf},  and Mobilenerf~\cite{chen2022mobilenerf},  aim to use various advanced technologies to speed up the training or inference phases. Though these methods have achieved great progress, due to the potential issues of inaccurate camera poses, simplified pinhole camera models, and scene representation inaccuracy, they still suffer from rendering artifacts when predicting novel views. 
	
	\vspace{0.03in}
	\noindent\textbf{Degradation simulation.}
	Since there are currently no attempts to explore NeRF-style degradation, we overview the real-world image restoration works that are most related to ours. The previous image and video super-resolution approaches~\cite{li2022best,dong2015image,zhang2018residual,zhou2022revisiting,wang2019edvr,li2020lapar,wang2018esrgan,liang2021swinir,zhang2018residual,yu2021path} typically follow a fixed image degradation type~(\textit{e.g.}, blur, bicubic or bilinear down-sampling). Due to the large domain shift between the real-world and simulated degradations, the earlier image restoration methods~\cite{li2020lapar,li2019feedback,zhang2018residual,zhang2020deep} generally fail to remove complex artifacts of the real-world images. In contrast, BSRGAN~\cite{zhang2021designing} designs a practical degradation approach for real-world image super-resolution. In their degradation process, multiple degradations are considered and applied in random orders, largely covering the diversity of real-world degradations. Compared with previous works, BSRGAN achieves much better results quantitatively and qualitatively. Real-ESRGAN~\cite{wang2021real} develops a second-order degradation process for real-world image super-resolution. Unlike the real-world image and video processing systems that focus on eliminating image and video compression, motion blur, video interlace, and sensor noise, the NeRF-rendering involves different degradation patterns. To the best knowledge, we are the first to investigate NeRF-style degradation removal.

	\vspace{0.03in}
	\noindent\textbf{Correspondence estimation.}
	In the existing literature, video restoration methods~\cite{yu2020joint,wang2018learning,teed2020raft,chan2021basicvsr++,cao2021vsrt} aim to restore a high-quality frame from multiple low-quality frames. To achieve this goal, cross-frame correspondence estimation is essential to effectively aggregate informative temporal contents. Some works~\cite{yu2020joint,xue2019video,chan2021basicvsr++,chan2021basicvsr} build pixel-level correspondences through optical-flow estimation and perform frame warping for multi-frame compensation. Another line of works~\cite{wang2019edvr,tian2020tdan,zhou2022revisiting} tries to use deformable convolution networks~(DCNs~\cite{dai2017deformable}) for adaptive correspondence estimation and aggregation. More recently, transformer-based video restoration models~\cite{liang2022vrt,cao2022vdtr} implement spatial-temporal aggregation through an attention mechanism and achieve promising performance. However, it is still challenging to perform accurate correspondence estimation between frames captured with very distinctive viewpoints.
	
	{\color{black}
		\noindent\textbf{High-resolution NeRFs (HR-NeRFs).}
		To enhance visual quality in high-resolution rendering, various NeRF techniques~\cite{wang2022nerf,wang20224k,yoon2023cross,huang2023refsr} have been developed. NeRF-SR~\cite{wang2022nerf} employs a two-stage training process, initially using super-sampling to generate a high-resolution output from low-resolution views and then refining it by incorporating information from reference patches. 4K-NeRF~\cite{wang20224k} adopts a similar approach, utilizing DVGO~\cite{Sun_2022_CVPR} to approximate the 3D scene representation and employing a view-consistent encoder-decoder for high-quality rendering. Ref-SR NeRF~\cite{huang2023refsr} optimizes a low-resolution NeRF representation and integrates a reference-based super-resolution model to construct high-resolution views.  CROP-NeRF~\cite{yoon2023cross} employs a cross-optimization scheme to improve the NeRF representation by simultaneously optimizing a deep image super-resolution model. 
		
		By comparison, our NeRFLiX/NeRFLiX++ offers superiority with its scene-agnostic training and cost-effective reconfiguration. First, our method distinguishes itself from existing approaches, which necessitate multi-stage scene-specific training, by employing a single training process. The direct application to new scenes substantially reduces the training overhead. Second, in traditional HR-NeRFs, the joint optimization of the initial NeRF model and the refinement model makes them tightly coupled, leading to potential performance degradation if either component is replaced. Conversely, our framework effectively decouples the NeRF rendering and refinement stages, enhancing adaptability to existing or future NeRFs. The Table~\ref{subtab:nerf-sr} and Table~\ref{table:generalization} results clearly demonstrate the substantial improvements, strong generalization capability, and improved computational efficiency achieved by NeRFLiX++.
	}
	\begin{table}[t]
		\small
		\setlength{\tabcolsep}{1pt}
		\begin{center}
			\begin{tabular}{l|c|c|c } 
				\hline
				Model		&56Leonard      &Transamerica    &Param./Time/Mem.        \\ \hline
				BungeeNeRF($\frac{1}{2}$)   &21.15/0.616 &21.50/0.585 & \textbf{ 0.3M}/11.04s/21.6GB \\ \hline
				\rowcolor{gray!15}	NeRF-SR 		&20.47/0.586         &20.96/0.551 &30.3M/2.73s/21.7GB                      \\ 
				NeRFLiX++$_{4K-L1}$		&\textbf{22.50/0.766}         &\textbf{22.83/0.741}         &14.4M/\textbf{0.43s}/\textbf{ 7.1GB}          \\ \hline
				
			\end{tabular}
		\end{center}
		\vspace{-0.1in}
		\caption{\color{black} Generalization analysis of NeRF-SR~\cite{wang2022nerf} and our proposed NeRFLiX++. Without re-training on the city-scale scenes~\cite{xiangli2022bungeenerf}, we directly evaluate their enhancement abilities using PSNR (dB)/SSIM metrics. The best results are highlighted in \textbf{bold}. Additionally, we assess the inference costs of all three models, calculated for a resolution of 1024$\times$1024 on an NVIDIA RTX 3090.}
		\label{table:generalization}
		\vspace{-0.15in}
	\end{table}
	\section{Preliminaries}
	\begin{figure}[ht]
		\centering
		\includegraphics[width=0.95\columnwidth]{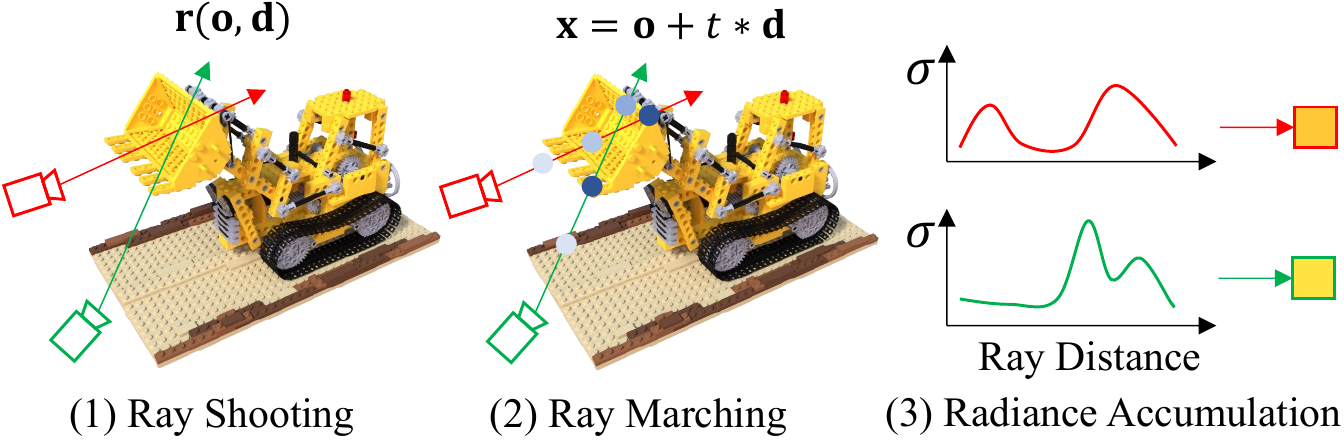} 
		\caption{A general illustration of NeRF-based novel view synthesis pipeline. Three main steps are involved: (1) ray shooting, (2) ray marching, and (3) radiance accumulation.
		} 
		\vspace{-0.1in}
		\label{fig:NeRFIll}
	\end{figure} %
	
	In this section, we review the general pipeline of NeRF-based novel view synthesis and discuss potential rendering artifacts. As shown in Fig.~\ref{fig:NeRFIll}, three main steps are involved in the rendering.
	(1) Ray shooting.
	To render the color of a target pixel in a particular view, NeRF utilizes the camera's calibrated parameters~$\pi$ to generate a ray $\mathbf{r(o,d)}$ through this pixel, where $\mathbf{o}$ and $\mathbf{d}$ are the camera center and the ray direction.
	(2) Ray marching. 
	A set of 3D points are sampled along the chosen ray as it moves across the 3D scene represented by neural radiance fields. NeRF encodes a 3D scene and predicts the colors and densities of these points.
	(3) Radiance accumulation.
	The pixel color is extracted by integrating the predicted radiance features of the sampled 3D points.
	
	\vspace{0.03in}
	\noindent\textbf{Discussion.} We see that establishing relationships between 2D photos and the corresponding 3D scene requires camera calibration. Unfortunately, it is very challenging to precisely calibrate camera poses, leading to noisy 3D sampling. Meanwhile, some previous works~\cite{yariv2020multiview,jeong2021self,wang2021nerf,zhang2022vmrf} also raise other concerns, including the non-linear pinhole camera model~\cite{jeong2021self} and shape-radiance ambiguity~\cite{zhang2020nerf++}. Due to these inherent limitations, as discussed in Sec.~\ref{sec:introduction}, NeRF models may synthesize unsatisfied novel test views.
	
	\begin{figure}[t]
		\centering
		\includegraphics[width=0.95\columnwidth]{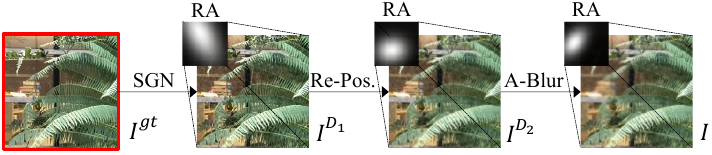} 
		
		\caption{Overview of our NDS pipeline in NeRFLiX: using our proposed degradations, we process a target view $I^{gt}$ to produce its simulated degraded view $I$. ``SGN", ``Re-Pos." and ``A-Blur" refer to the splatted Gaussian, re-positioning, anisotropic blur degradations, and ``RA" is the region adaptive strategy.
			\vspace{-0.15in}
		} 
		
		\label{fig:nds}
	\end{figure} %

	\begin{figure*}[ht]
		\centering
		\includegraphics[width=18cm]{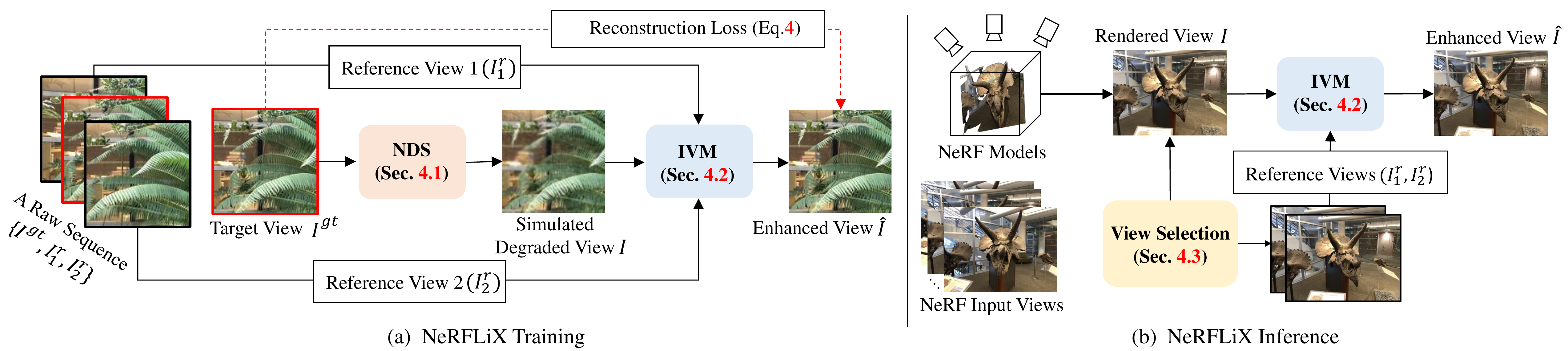} 
		\caption{Illustration of our proposed NeRFLiX. It consists of two essential modules: (1) NeRF degradation simulator that constructs paired training data $\{I,I_1^r,I_2^r|I^{gt}\}$ from a raw sequence $\{I^{gt},I_1^r,I_2^r\}$, (2) inter-viewpoint mixer trained on this simulated data is capable of restoring high-quality frames from NeRF rendered views. }
		\vspace{-0.18in}
		\label{fig:pipeline}
	\end{figure*} %
	
	\section{NeRFLiX}
	
	In this work, we present NeRFLiX, a general NeRF-agnostic restorer which employs a degradation-driven inter-viewpoint mixer to enhance novel view images rendered by NeRF models.
	It is made up of two essential components: a NeRF-style degradation simulator~(NDS) and an inter-viewpoint mixer~(IVM). As shown in Fig.~\ref{fig:pipeline}{\color{red}a}, during the training phase, we employ the proposed NDS to create large-scale paired training data, which is subsequently used to train an IVM for improving a NeRF-rendered view utilizing two reference images (reference views). In the inference stage, as illustrated in Fig.~\ref{fig:pipeline}{\color{red}b}, IVM is adopted to enhance a rendered view by fusing useful information from the selected most relevant reference views.

	\vspace{-0.03in}
	\subsection{NeRF-Style Degradation Simulator~(NDS)}
	\label{sec:degraded}
	
	
	Due to the difficulties in gathering well-posed scenes under various environments and training NeRF models for each scene, it is infeasible to directly collect large amounts of \textit{paired} NeRF data for artifact removal.
	To address this challenge, motivated by BSRGAN~\cite{zhang2021designing},
	we design a general NeRF degradation simulator to produce a sizable training dataset that is visually and statistically comparable to NeRF-rendered images~(views).
	
	To begin with, we collect raw data from LLFF-T\footnote{The training part of LLFF~\cite{mildenhall2019local}.} and Vimeo90K\cite{xue2019video} where the adjacent frames are treated as raw sequences. Each raw sequence consists of three images $\{I^{gt},I_1^r,I_2^r\}$: a target view $I^{gt}$ and its two reference views $\{I_1^r,I_2^r\}$. To construct the paired data from a raw sequence, we use the proposed NDS to degrade $I^{gt}$ and obtain a simulated view $I$, as shown in Fig.~\ref{fig:pipeline}(a). 
	
	The degradation pipeline is illustrated in Fig~\ref{fig:nds}. We design three types of degradation for processing a target view $I^{gt}$: splatted Gaussian noise~(SGN), re-positioning~(Re-Pos.), and anisotropic blur~(A-Blur). It should be noted that \textit{there may be other models for such a simulation}, and we only utilize this route to evaluate and justify the feasibility of our idea.
	
	\vspace{0.03in}
	\noindent\textbf{Splatted Gaussian noise.}
	Although additive Gaussian noise is frequently employed in image and video denoising, NeRF rendering noise clearly differs. Rays that hit a 3D point will be re-projected within a nearby 2D area because of noisy camera parameters. As a result, the NeRF-style noise is dispersed over a 2D space. This observation led us to present a splatted Gaussian noise, which is defined as
	\begin{equation}
		I^{D1} = (I^{gt}  + n) \circledast g \,,
		\vspace{-0.05in}
	\end{equation}
	where $n$ is a 2D Gaussian noise map with the same resolution as $I^{gt}$ and $g$ is an isotropic Gaussian blur kernel.

	\vspace{0.03in}
	\noindent\textbf{Re-positioning.}
	We design a re-positioning degradation to simulate ray jittering. We add a random 2D offset $\delta_i,\delta_j \in [-2,2]$ with probability 0.1 for a pixel at location $(i,j)$ as
	\vspace{-0.05in}
	\begin{equation}
		I^{D2}(i,j) = 
		\begin{cases}
			I^{D1}(i,j) &\text{if~~} p>0.1 \\
			I^{D1}(i+ \delta_i,j+\delta_j) &\text{else~~} p\leq0.1 \,,
		\end{cases}
		\vspace{-0.05in}
	\end{equation}
	where $p$ is uniformly distributed in $[0, 1]$.

	\vspace{0.03in}
	\noindent\textbf{Anisotropic blur.}
	From our observation, NeRF synthetic frames also contain blurry contents. To simulate blur patterns, we use anisotropic Gaussian kernels to blur the target frame.

	
	
	\vspace{0.03in}
	Neural radiance fields are often supervised with unbalanced training views. As a result, given a novel view, the projected 2D areas have varying degradation levels. Thus, we carry out each of the employed degradations in a spatially variant manner. More specifically, we define a mask $M$ as a two-dimensional oriented anisotropic Gaussian~\cite{geusebroek2003fast} like
	\begin{equation} 
		\vspace{-0.02in}
		M(i,j)= G(i-c_i, j-c_j; \sigma_i, \sigma_j, A)\,,
		\vspace{-0.02in}
	\end{equation} 
	where $(c_i,c_j)$ and $(\sigma_i, \sigma_j)$ are the means and standard deviations, and $A$ is an orientation angle. After that, we use the mask $M$ to linearly blend the input and output of each degradation, finally achieving region-adaptive degradations. 
	
	At last, with our NDS, we can obtain a great number of training pairs, and each paired data consists of two high-quality reference views $\{I_1^{r},I_2^{r}\}$, a simulated degraded view $I$, and the corresponding target view $I^{gt}$. Next, we show how the constructed paired data $\{I,I_1^{r},I_2^{r}|I^{gt}\}$ can be used to train our IVM.

	
	\begin{figure}[t]
		\centering
		\includegraphics[width=1.0\columnwidth]{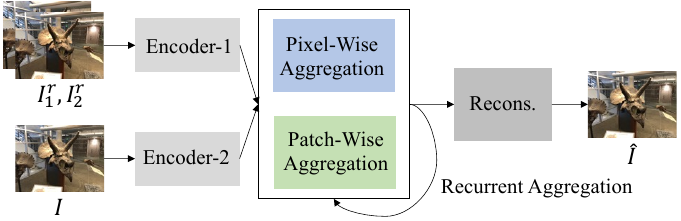} 
		
		\caption{The framework of our inter-viewpoint mixer in NeRFLiX.
		} 
		\vspace{-0.1in}
		\label{fig:ivm}
	\end{figure} %
	
	\subsection{Inter-viewpoint Mixer~(IVM)}
	\label{sec:ivm}
	
	\noindent\textbf{Problem formulation.} Given a degraded view $I$ produced by our NDS or NeRF models, we aim to extract useful information from its two high-quality reference views $\{I_1^r,I_2^r\}$ and restore an enhanced version $\hat I$.

	\vspace{0.03in}
	\noindent\textbf{IVM architecture.}
	For multi-frame processing, the existing techniques either use optical flow~\cite{chan2021basicvsr,yu2020joint,wang2018learning} or deformable convolutions~\cite{dai2017deformable,wang2019edvr,liang2022vrt} to realize the correspondence estimation and aggregation for \textit{consistent} displacements. In contrast, NeRF-rendered and input views may come from very different angles and locations, making it challenging to perform precise aggregation.
	
	To address this problem, we propose IVM, a hybrid recurrent inter-viewpoint ``mixer" that progressively fuses pixel-wise and patch-wise contents from two high-quality reference views, achieving more effective inter-viewpoint aggregation. There are three modules, \textit{i.e.}, feature extraction, hybrid inter-viewpoint aggregation, and reconstruction, as shown in Fig.~\ref{fig:ivm}. Two convolutional encoders are used in the feature extraction stage to process the degraded view $I$ and two high-quality reference views $\{I_1^r,I_2^r\}$, respectively. We then use inter-viewpoint window-based attention modules and deformable convolutions to achieve recurrent patch-wise and pixel-wise aggregation. Finally, the enhanced view ${\hat I}$ is generated using the reconstruction module under the supervision
	\vspace{-0.05in}
	\begin{equation}
		\begin{split}
			Loss = |{\hat I} - I^{gt}|, \text{where  } {\hat I} = f(I,I_1^r,I_2^r;\theta)\,,
		\end{split}
		\vspace{-0.05in}
	\end{equation}
	where $\theta$ is the learnable parameters of IVM. The architecture details are given in our supplementary materials.
	
	\begin{figure}[t]
		\centering
		\includegraphics[width=0.95\columnwidth]{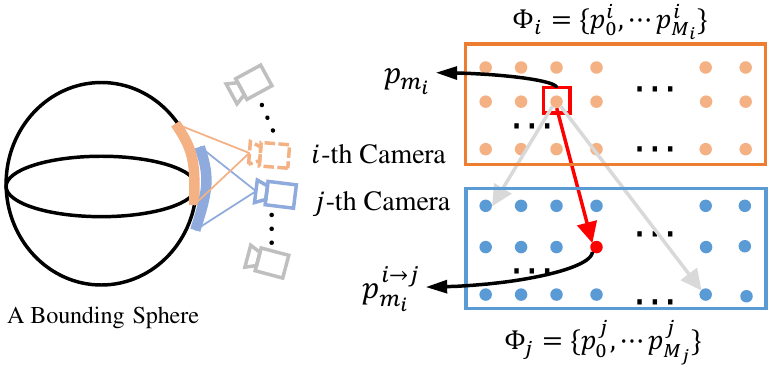} 
		
		\caption{Illustration of our view selection strategy. 
		} 
		\vspace{-0.1in}
		\label{fig:vm}
	\end{figure} %
	
	\subsection{View Selection}
	
	\begin{figure*}[t]
		\begin{center}
			\includegraphics[width=\linewidth]{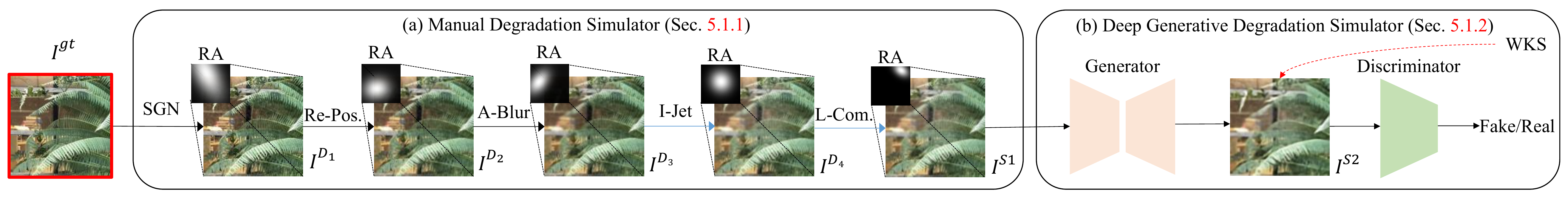} 
		\end{center}
		\vspace{-0.1in}
		\caption{The pipeline of our two-stage degradation modeling consisted of two sequentially stacked simulators: (a) a manual degradation simulator to get an initialized result $I^{S1}$ from a clean target frame, (b) a deep generative degradation simulator that receives the $I^{S1}$ and outputs the final degraded frame $I^{S2}$ using an adversarial learning scheme. Additionally, we introduce a weighted top-$K$ supervision~(WKS) to enhance the degradation diversity of refined views $I^{S2}$.
			\vspace{-0.1in}
		}
		\label{fig:deepNDS}
	\end{figure*}
	
	In the inference stage, for a NeRF-rendered view $I$, our IVM produces an enhanced version by aggregating contents from two neighboring high-quality views. Though multiple high-quality views (provided for the training) are available, only a part of them is largely overlapped with $I$. We only adopt the most pertinent views that are useful for the inter-viewpoint aggregation.
	
	To this end, we develop a view selection strategy to choose two reference views $\{I_1^r,I_2^r\}$ from the input views that are most overlapped with the rendered view $I$. Specifically, we formulate the view selection problem based on the pinhole camera model. An arbitrary 3D scene can be roughly approximated as a bounding sphere in Fig.~\ref{fig:vm}, and cameras are placed around it to take pictures. When camera-emitted rays hit the sphere, there are a set of intersections. We refer to the 3D point sets as $\Phi_i=\{p_{0}^i,p_{1}^i,\cdots,p_{M_i}^i\}$ and $\Phi_j=\{p_{0}^j,p_{1}^j,\cdots,p_{M_j}^j\}$ for the $i$-th and $j$-th cameras. For $m_i$-th intersection $p_{m_i}^i \in \Phi_i$ of view $i$, we search its nearest point in view $j$ with the L2 distance
	\begin{equation} 
		p_{m_i}^{i \rightarrow j} = \mathop{\arg\min}_{p \in \Phi_j} ( || p - p_{m_i}^i||_2^2 )\,.
	\end{equation}
	Then the matching cost from the $i$-th view to the $j$-th view is calculated by
	\vspace{-0.1in}
	\begin{equation}
		C_{i \rightarrow j} = \sum_{m_i=0}^{M_i}|| p_{m_i}^{i} - p_{m_i}^{i \rightarrow j}||_2^2\,.
	\end{equation}
	We finally obtain the mutual matching cost between views $i$ and $j$
	\begin{equation}
		C_{i \leftrightarrow j} = C_{i \rightarrow j} + C_{j \rightarrow i}\,.
		\label{eq:mutualcost}
	\end{equation}
	In this regard, two reference views $\{I_1^r,I_2^r\}$ are selected at the least mutual matching costs for enhancing the NeRF-rendered view $I$. Note that we also adopt this strategy to decide the two reference views for the LLFF-T~\cite{mildenhall2019local} data during the training phase.

	\section{NeRFLiX++}
	Based on NeRFLiX, we propose NeRFLiX++ with a two-stage degradation modeling strategy and a guided inter-viewpoint mixer to further improve restoration performance and efficiency.
	
	
	\subsection{Two-stage Degradation Modeling}
	The proposed two-stage degradation modeling approach comprises a manually designed degradation simulator and a deep generative degradation simulator, as depicted in Fig.~\ref{fig:deepNDS}. In the first stage, we generate initialized degraded frames using multiple hand-crafted degradations, inspired by NeRFLiX, from the selected clean views. In the second stage, the deep generative degradation simulator is employed to refine the first-stage results and generate the final simulated views.
	
	\begin{figure}[t]
		\centering
		\includegraphics[width=0.9\columnwidth]{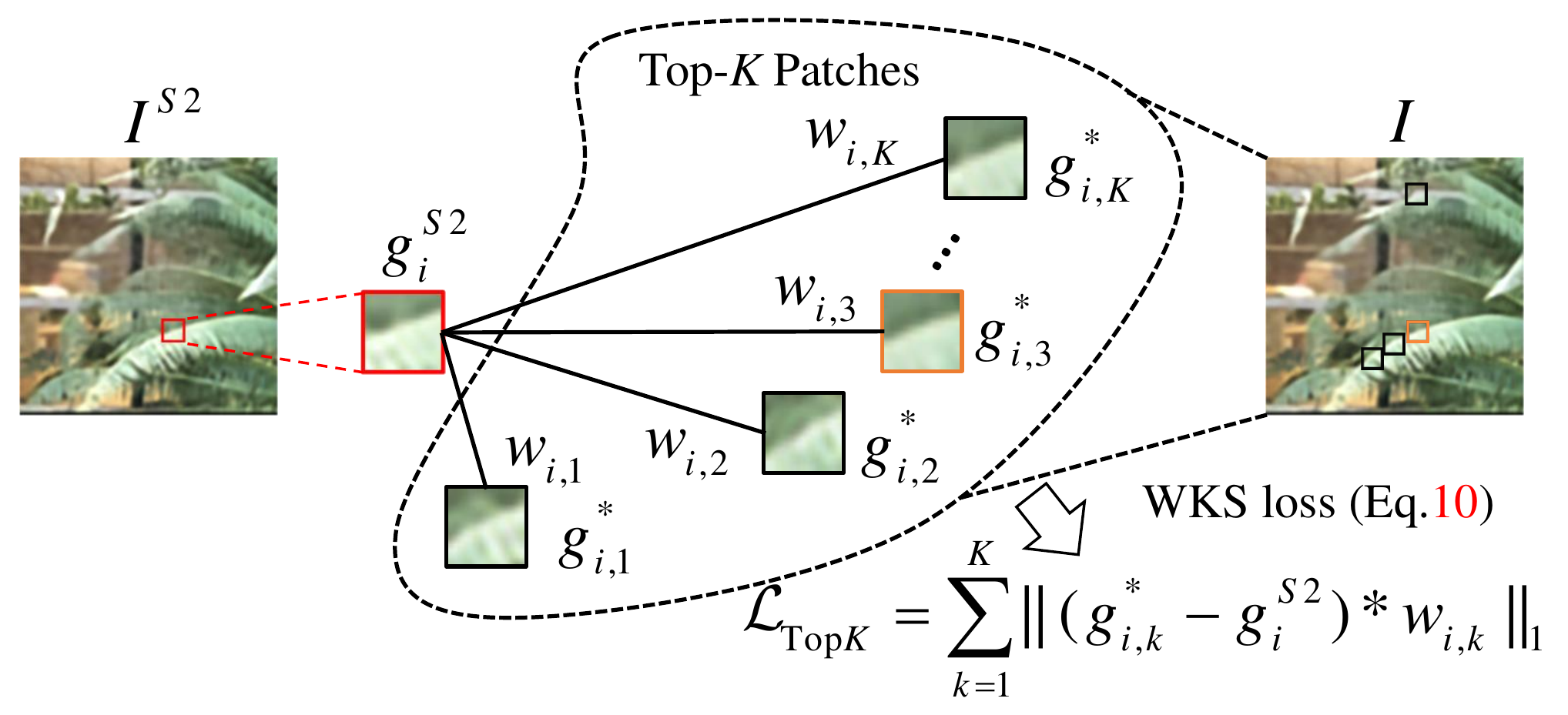} 
		
		\caption{Illustration of the proposed weighted top-$K$ similarity loss. Where $I^{S2}$ and $I$ are the second-stage degraded frame and the corresponding NeRF-rendered views. $\mathbf{g}_{i, \{1,2,\cdots,K\}}^*$ are the top-$K$ patches from $I$ and $w_{i, \{1,2,\cdots,K\}}$ are the corresponding weighted factors calculated by Eq.~(\ref{eq:w}) for adaptive supervision. Also, we highlight the pre-defined patch $\mathbf{g}_i^{gt}$ that is only ranked third.
		} 
		\label{fig:WKS}
		\vspace{-0.15in}
	\end{figure} %
	\subsubsection{Manual Degradation Simulator}
	
	In addition to the three basic degradations used in NeRFLiX, which are splatted Gaussian noise, re-positioning, and anisotropic Gaussian blur, we introduce two supplementary degradation patterns to enhance the realism of our simulation. We apply the same region-adaptive degradation strategy as NeRFLiX for these two additional degradations.
	
	\vspace{0.03in}
	\noindent\textbf{Illumination jetting.}
	To account for the variation in illumination caused by view-dependent shading, we propose a gamma adjustment applied to \textit{both} the target and reference views. The adjustment is defined as
	\begin{equation}
		y = \mathrm{power}(x,\gamma)\,,
	\end{equation}
	where ``$\mathrm{power}$" denotes the exponential function and $\gamma$ is a linear adjustment constant randomly sampled from $[0.95,1.05]$.
	
	\vspace{0.05in}
	\noindent\textbf{Lightness compression.} To simulate structural defects that may occur in NeRF-based rendering, we propose an image compression procedure that degrades the gray-scale density of a target frame. Specifically, we first convert an RGB frame to the LAB color space and compress the L component using the JPEG algorithm at a randomly selected compression level (between 20$\%$ and 90$\%$). We then merge the degraded L channel with the raw AB channels and transform them back to the RGB color space. 
	
	\begin{figure*}[t]
		\begin{center}
			\includegraphics[width=\linewidth]{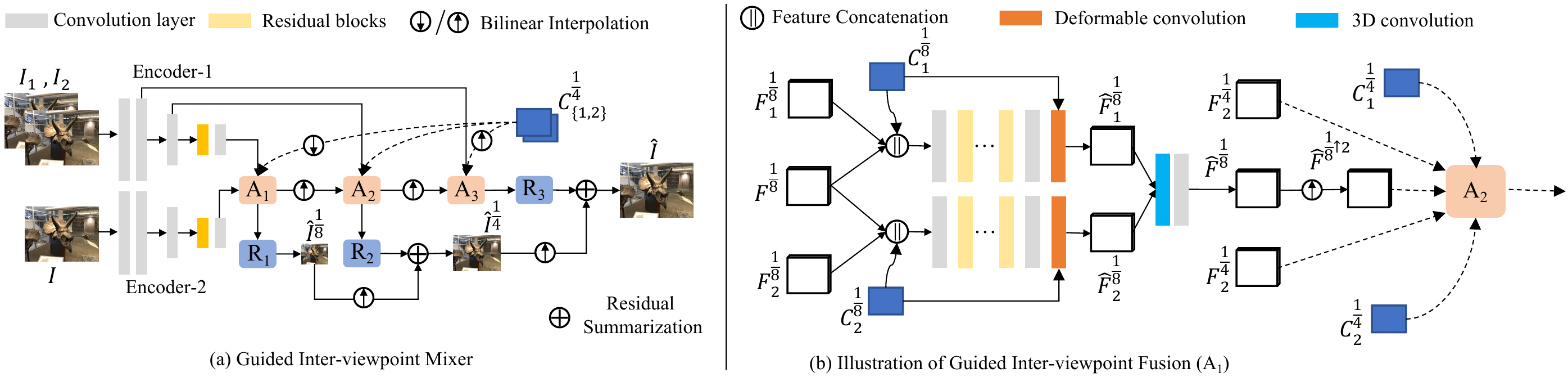} 
		\end{center}
		\vspace{-0.15in}
		\caption{The framework of our guided inter-viewpoint mixer (G-IVM) in NeRFLiX++. (a) The overview of our proposed G-IVM, which consists of three integral modules: (1) two parallel convolutional encoders are employed to extract deep image features from input and reference views, (2) taking the estimated coarse optical flow as guidance, we devise a guided inter-viewpoint aggregation module to progressively fuse pyramid deep image features, and (3) we utilize three reconstruction blocks to gradually restore multi-scale frames in a coarse-to-fine manner. Meanwhile, we adopt a multi-scale supervision scheme to enhance the restoration quality. (b) We outline the detailed structure of the guided inter-viewpoint aggregation. ``A$_{\{1,2,3\}}$" means the three pyramid aggregation blocks, and ``R$_{\{1,2,3\}}$" refer to the three reconstruction blocks to progressively produce multi-scale frames ${\hat I}^{\{\frac{1}{8},\frac{1}{4}\}}$ and ${\hat I}$.
		}
		\vspace{-0.05in}
		\label{fig:GIVMF}
	\end{figure*}

	\subsubsection{Deep Generative Degradation Simulator}\label{sec:dgds}
	
	As noted in Sec.~\ref{sec:introduction}, manually designed degradations may not capture the full range of actual NeRF-style artifacts. To address this limitation, we propose a deep generative degradation simulator that refines the results of the manual degradation stage and narrows the gap between the simulated and target domains.

	
	Generative adversarial networks (GANs) have shown remarkable results in image-to-image translation tasks when a large number of training samples are available. However, the scarcity of available data from Neural Radiance Fields (NeRF) poses significant challenges in using GANs to directly fit the underlying degradation distribution. To address this issue, motivated by Beby-GAN~\cite{li2022best}, we propose a weighted top-$K$ similarity loss, or WKS, as an auxiliary loss function to aid in adversarial training. As shown in Fig.~\ref{fig:deepNDS}, we use a UNet to process the first-stage degraded 
	view $I^{S1}$ to obtain the refined result $I^{S2}$. In addition to the conventional adversarial and reconstruction losses, we also utilize the WKS to produce results $I^{S2}$ with more diversity.
	
	\vspace{0.03in}
	\noindent\textbf{WKS.} Fig.~\ref{fig:WKS} illustrates the weighted top-$K$ supervision. Given the $i$-th patch of $I^{S2}$, denoting $\mathbf{g}^{S2}_{i}$, we use a triple distance function to search for the top-$K$ similar patches $\mathbf{g}_{i,\{1,2,\cdots, K\}}^{*}$ from the corresponding real rendered view $I$. The triple distance function is defined as
	\begin{equation}
		\begin{aligned}
			\mathbf{g}_{i,\{1,2,\cdots, K\}}^{*} = \underset{\mathbf{g} \in \mathcal{G}}{\rm topK} \ \alpha {\left\| \mathbf{g} - \mathbf{g}^{S2}_{i}\right\|}_{2}^{2} + \beta {\left\| \mathbf{g} - \mathbf{g}^{gt}_{i} \right\|}_{2}^{2}\,,
		\end{aligned}
		\label{eq:bbf}
	\end{equation}
	where $\mathbf{g}^{gt}_{i}$ is the corresponding real rendered patch, $\mathcal{G}$ is a set of candidate patches generated by unfolding the real-rendered view $I$, and $\alpha,\beta$ are two scaling factors to balance the two distance terms. According to the empirical experiments in Beby-GAN~\cite{li2022best}, we set them to 1 for better evaluation results. After obtaining the top-$K$ similar patches, the proposed WKS is formulated as
	\begin{equation}
		\mathcal{L}_{\rm TopK} = \sum_{k=1}^{K} \left\| (\mathbf{g}_{i, k}^{*} - \mathbf{g}^{S2}_{i}) * w_{i, k} \right\|_{1} ,
		\label{eq:topk}
	\end{equation}
	where $w_{i, k}$ is the $k$-th normalized weight, calculated as
	\begin{equation}
		\begin{split}
			d_{i, k} &= -\frac{1}{2}||\mathbf{g}_{i, k}^{*} -  \mathbf{g}^{S2}_{i}||_2^2\,,\\
			w_{i, k} &= \mathop{\rm  exp}(d_{i, k}) / \sum_{m=1}^K\mathop{\rm exp}(d_{i, m})\,.
		\end{split}
		\label{eq:w}
	\end{equation}
	The $d_{i, k}$ is the scaled negative L2 distance between the predicted patch $\mathbf{g}^{S2}_{i}$ and one of its $k$-th most similar patch.

	\vspace{0.03in}
	\noindent\textbf{Discussion.} Our proposed weighted top-$K$ similarity loss adopts a dynamic strategy to search for multiple pertinent patches from the real rendered frames, enriching the diversity of supervisory signals. This approach encourages the model to find highly similar target patches that have closer degradation degrees than the pre-defined label, resulting in more accurate and effective training. In our experiments, we demonstrate the effectiveness of this design and show that it significantly improves the performance of GANs when limited data from NeRF-rendered frames is available.
	
	
	\begin{table*}[t]
		\label{tab:nsds}
		\small
		\setlength{\tabcolsep}{3pt}
		\begin{subtable}[b]{0.46\textwidth}
			
			\begin{tabular}{l|c|c|c} 
				\hline
				Method                      & PSNR~(dB)${\color{red}\uparrow}$ &SSIM${\color{red}\uparrow}$ &LPIPS${\color{red}\downarrow}$   \\ \hline
				
				TensoRF~\cite{tensorf}  {\footnotesize (ECCV'22)}          			 &  26.73      		     & 0.839  &0.204  \\ 
				TensoRF~\cite{tensorf} + NeRFLiX                 & {\color{red}27.39} ($\color{red}\uparrow$ 0.66)         &  {\color{red}0.867} &{\color{red}0.149} \\ 
				TensoRF~\cite{tensorf} + NeRFLiX++                 &  {\color{blue}27.38} ($\color{red}\uparrow$ 0.65)         &  {\color{blue}0.866} &{\color{blue}0.156} \\ 
				\hline \hline

				Plenoxels~\cite{fridovich2022plenoxels}  {\footnotesize(CVPR'22)}            			 &  26.29      		     & 0.839  &0.210  \\ 
				Plenoxels~\cite{fridovich2022plenoxels} + NeRFLiX                 &  {\color{blue}26.90} ($\color{red}\uparrow$ 0.61)         &  {\color{red}0.864} &{\color{red}0.156}  \\  
				Plenoxels~\cite{fridovich2022plenoxels} + NeRFLiX++                 &  {\color{red}26.92} ($\color{red}\uparrow$ 0.63)         &  {\color{red}0.864} &{\color{blue}0.160}  \\\hline \hline

				NeRF-mm~\cite{wang2021nerf}  {\footnotesize(ARXIV'21) }            			 & 22.98    		     & 0.655  &0.440  \\ 
				NeRF-mm~\cite{wang2021nerf} + NeRFLiX                 &  {\color{blue}23.38} ($\color{red}\uparrow$ 0.40)         &  {\color{blue}0.694} &{\color{blue}0.360} \\
				NeRF-mm~\cite{wang2021nerf} + NeRFLiX++                 & {\color{red}23.40} ($\color{red}\uparrow$ 0.42)         &  {\color{red}0.698} &{\color{red}0.354} \\
				\hline \hline
				
				NeRF~\cite{mildenhall2020nerf} {\footnotesize (ECCV'20)}           			 & 26.50    		     & 0.811  &0.250  \\ 
				NeRF~\cite{mildenhall2020nerf} + NeRFLiX                  &{\color{red}27.26} ($\color{red}\uparrow$ 0.76)         &  {\color{red}0.863} &{\color{red}0.159}  \\ 
				NeRF~\cite{mildenhall2020nerf} + NeRFLiX++                 &{\color{blue}27.25} ($\color{red}\uparrow$ 0.75)         &   {\color{blue}0.858} &{\color{blue}0.170} \\
				\hline
			\end{tabular}
			\caption{Quantitative results on LLFF~\cite{mildenhall2019local} under LLFF-P1.}
			\label{subtab:sotallff1}
		\end{subtable}
		\hfill
		\begin{subtable}[b]{0.46\textwidth}
			
			\begin{tabular}{l|c|c|c } 
				\hline
				Method                      & PSNR~(dB)${\color{red}\uparrow}$ &SSIM${\color{red}\uparrow}$ &LPIPS${\color{red}\downarrow}$   \\ \hline
				
				NLF~\cite{attal2022learning} {\footnotesize(CVPR'22)}           			 & 27.46     		     & 0.868  &0.136  \\ 
				NLF~\cite{attal2022learning} + NeRFLiX                 &  {\color{red}28.19} ($\color{red}\uparrow$ 0.73)         &   {\color{red}0.899} &{\color{red}0.093} \\  
				NLF~\cite{attal2022learning} + NeRFLiX++                 &  {\color{blue}28.10} ($\color{red}\uparrow$ 0.64)         &   {\color{blue}0.895} &{\color{red}0.093} \\  \hline \hline
				RegNeRF-V3~\cite{Niemeyer2021Regnerf} {\footnotesize(CVPR'22)}    & 19.10		     & 0.587  &0.373  \\ 
				RegNeRF-V3~\cite{Niemeyer2021Regnerf} + NeRFLiX    &{\color{blue}19.68} ($\color{red}\uparrow$ 0.58)		     &{\color{blue} 0.661}  &{\color{blue}0.260} \\
				RegNeRF-V3~\cite{Niemeyer2021Regnerf} + NeRFLiX++    &{\color{red}19.85} ($\color{red}\uparrow$ 0.75)		     &{\color{red} 0.670}  &{\color{red}0.258}  \\ \hline
				
				RegNeRF-V6~\cite{Niemeyer2021Regnerf} {\footnotesize(CVPR'22)}    &23.06		     & 0.759  &0.242  \\ 
				RegNeRF-V6~\cite{Niemeyer2021Regnerf} + NeRFLiX    &{\color{blue}23.90}($\color{red}\uparrow$ 0.84)		     & {\color{blue}0.815}  &{\color{red}0.144}  \\
				RegNeRF-V6~\cite{Niemeyer2021Regnerf} + NeRFLiX++    &{\color{red}24.01}($\color{red}\uparrow$ 0.95)		     & {\color{red}0.816}  &{\color{blue}0.152}  \\\hline
				
				RegNeRF-V9~\cite{Niemeyer2021Regnerf} {\footnotesize(CVPR'22)}    & 24.81     		     & 0.818  &0.196  \\ 
				RegNeRF-V9~\cite{Niemeyer2021Regnerf} + NeRFLiX    &{\color{blue}25.68} ($\color{red}\uparrow$ 0.87)		     &{\color{red} 0.863}  &{\color{red}0.114 } \\
				RegNeRF-V9~\cite{Niemeyer2021Regnerf} + NeRFLiX++    &{\color{red}25.76} ($\color{red}\uparrow$ 0.95)		     &{\color{blue} 0.861}  &{\color{blue}0.124 } \\\hline
			\end{tabular}
			
			\caption{Quantitative results on LLFF under LLFF-P2. RegNeRF-V3(6,9) takes 3(6,9) input views for training.}
			\label{subtab:sotallff2}
		\end{subtable}
		\hfill
		\vspace{0.1in}
		\begin{subtable}[b]{0.97\linewidth}
			\centering
			\begin{tabular}{l|c|cc|cc|cc} 
				\hline 
				\multirow{2}{*}{Model} & \multirow{2}{*}{$\#$Params} & \multicolumn{2}{c|}{@ $512\times512$} & \multicolumn{2}{c|}{@$1024\times 1024$} & \multicolumn{2}{c}{@$2048\times 2048$}\\ 
				& &\quad  Speed \quad &\quad  Memory \quad & 
				\quad Speed \quad & \quad Memory \quad & \quad Speed \quad & \quad Memory \quad \\
				\hline
				NeRF~\cite{mildenhall2020nerf} &5.1M  &14.01s & 22.5GB  &56.04s & 22.5GB  & 224.14s & 22.5GB \\
				\rowcolor{gray!15}	NeRF-mm~\cite{wang2021nerf}  &0.16M  &1.11s & 24.1GB  &4.45s & 24.1GB  & 17.80s & 24.1GB \\
				NLF~\cite{attal2022learning}     &1.3M   &4.19s & 15.8GB  &17.18s & 15.8GB  & 69.14s & 15.8GB \\ 
				\rowcolor{gray!15}	Reg-NeRF~\cite{Niemeyer2021Regnerf}  &0.6M   &7.54s & 23.8GB  &30.16s & 23.8GB  &120.65s & 23.8GB \\ 
				\rowcolor{gray!15}		Plenoxels~\cite{fridovich2022plenoxels}     &778M   &0.08s & 6.9GB  &0.29s & 6.9GB  & 1.16s & 6.9GB \\
				TensoRF~\cite{tensorf}     &47M   &4.15s & 21.6GB  &16.32s & 21.6GB  & 64.44s & 21.6GB \\ \hline \hline 
				NeRFLiX &35.2M     &0.92s & 4.2GB  &4.01s & 12.6GB  & - & -\\
				NeRFLiX++ &14.4M     &0.11s & 2.8GB   &0.43s & 7.1GB &1.58s & 24.2GB \\ \hline
				
			\end{tabular}
			\caption{\color{black} 
				{Analysis of model complexity and efficiency. Testing is conducted on an NVIDIA RTX 3090 GPU for three resolutions: $512\times 512$, $1024\times 1024$, and $2048\times 2048$. A hyphen ("-") denotes results that are unavailable due to out-of-memory constraints.}
			}
			\label{subtab:modelinference}
		\end{subtable}

		\caption{Quantitative analysis of our NeRFLiX on LLFF~\cite{mildenhall2019local}. Best and second best results are highlighted in {\color{red}red} and  {\color{blue}blue}.} \vspace{-0.1in}
		\label{tab:sotallff}
		\label{tab:sota1}
	\end{table*}
	
	\begin{figure*}[h]
		\begin{center}
			\includegraphics[width=\linewidth]{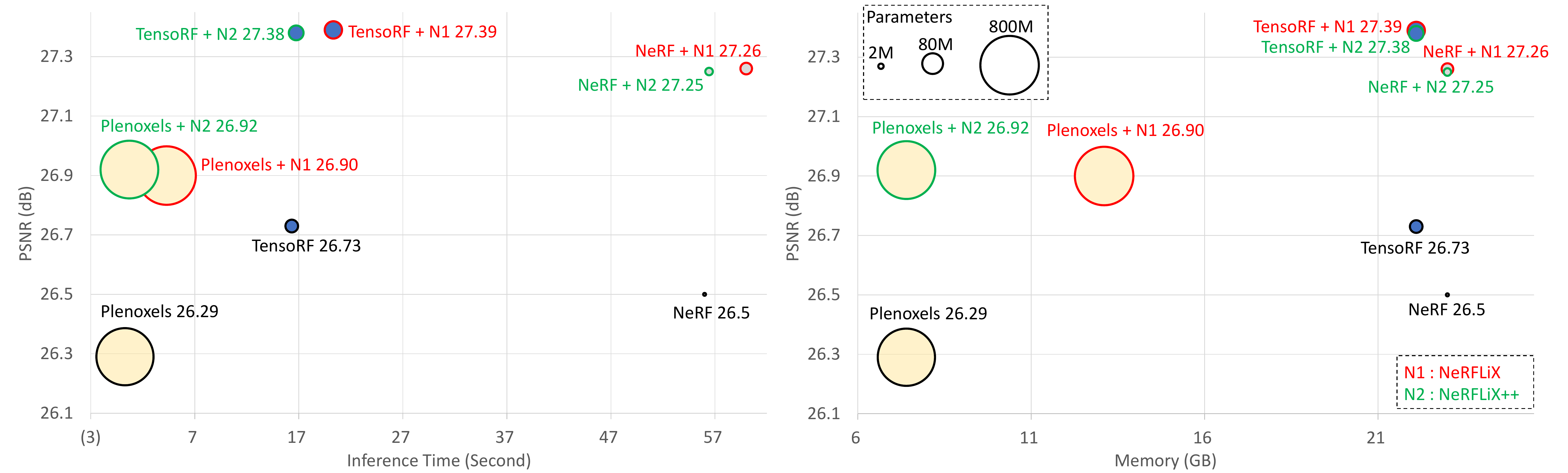} 
		\end{center}
		\vspace{-0.15in}
		\caption{\color{black}
			{Comparison between NeRF models and our NeRFLiX(++) additions in terms of PSNR, inference time, and memory usage. The runtime of NeRFLiX(++) is computed by aggregating the inference time of the NeRF model and IVM (G-IVM). Memory usage is reported as the maximum value between the NeRF model and IVM (G-IVM).}
		}
		\label{fig:speedcom}
		\vspace{-0.1in}
	\end{figure*}

	\subsection{Guided Inter-viewpoint Mixer}
	Under the typical NeRF setup, the high-quality input views come for free and they serve as potential reference bases for the restoration of rendered views.
	To achieve inter-viewpoint mixing, NeRFLiX presents a recurrent aggregation model to handle the distinct viewpoint changes. However, as aforementioned in Sec.~\ref{sec:introduction}, it remains impractical to process high-resolution frames due to the high computational expenses.
	
	To overcome this limitation, we propose a guided inter-viewpoint mixer~(termed as ``G-IVM") with an efficient multi-view fusion module.  Fig.~\ref{fig:GIVMF} depicts the framework architecture of G-IVM. Our approach first utilizes an off-the-shell optical flow model to predict coarse correspondences between a rendered view $I$ and its reference views $\{I_1, I_2\}$\footnote{To provide a more concise description of G-IVM, we omit the upper letter $^r$ in this section and the subsequent ones.} at a low resolution. Building upon coarse predictions as guidance, we propose a pyramid neural network to conduct a coarse-to-fine aggregation.
	
	Our guided inter-viewpoint mixer, an extension of the IVM method introduced in NeRFLiX~\cite{Zhou_2023_CVPR}, comprises three integral modules: feature extraction, guided inter-viewpoint aggregation, and pyramid reconstruction.
	
	\vspace{0.03in}
	\noindent\textbf{Coarse corresponding estimation.}  In order to establish coarse correspondences between a given input view $I$ and its reference views $\{I_1, I_2\}$, we utilize a pre-trained SPyNet~\cite{ranjan2017optical} model to predict the optical flow $C_{\{1,2\}}^{\frac{1}{4}}$ at a down-sampled scale of $\frac{1}{4}$.
	
	\vspace{0.03in}
	\noindent\textbf{Feature extraction.} We introduce two convolutional encoders, referred to as "Encoder-1/2", to extract deep pyramid image features $F^{\{\frac{1}{8},\frac{1}{4},\frac{1}{2}\}}$ and $F^{\{\frac{1}{8},\frac{1}{4},\frac{1}{2}\}}_{\{1,2\}}$ from a rendered view $I$ and its two reference views $I_{\{1,2\}}$. Specifically, the pyramid features are at scales of $\frac{1}{8},\frac{1}{4},\frac{1}{2}$ by applying three convolutions with a stride length of 2.
	
	\vspace{0.03in}
	\noindent\textbf{Guided inter-viewpoint aggregation.} 
	Considering the difficulties associated with accurately estimating large displacements between a rendered view $I$ and its reference views $\{I_1, I_2\}$, we present a guided inter-viewpoint aggregation method that operates in a coarse-to-fine manner. Our approach employs the flow-guided deformable convolution (FDCN) technique, leveraging optical flow computed by the SPyNet to facilitate the aggregation of $F^{\frac{1}{8}}$ and its corresponding reference views ${F^{\frac{1}{8}}_1, F^{\frac{1}{8}}_2}$. 
	The process is formulated as

	\begin{equation}
		\begin{aligned}
			& C_i^{\frac{1}{8}} = \frac{1}{2}{\rm Bilinear}(C_i^{\frac{1}{4}}, \frac{1}{2})\,, \\
			& M^{\frac{1}{8}}_{i} = [F^{\frac{1}{8}},F^{\frac{1}{8}}_i,C_i^{\frac{1}{8}}]\,, \\
			& \hat{F}^{\frac{1}{8}}_i = \mathop{\rm FDCN}(F^{\frac{1}{8}}_i,M^{\frac{1}{8}}_{i},C_i^{\frac{1}{8}}), \\
		\end{aligned}
	\end{equation}
	where $i \in \{1,2\}$ is the reference index, ${\rm Bilinear}(\cdot,s)$ is a bilinear interpolation function~($s$ is the scaling factor), $M^{\frac{1}{8}}_{i}$ is an offset feature, and $\hat{F}^{\frac{1}{8}}_i$ denotes an aligned feature from the $i$-th reference view to the target image. Having obtained the two aggregated features, denoted as $\hat{F}^{\frac{1}{8}}_{\{1,2\}}$, we employ a 3D convolution layer to fuse them with the target-view feature $F^{\frac{1}{8}}$:
	\begin{equation}
		\hat{F}^{\frac{1}{8}} = \mathop{\rm Conv3D}(\hat{F}^{\frac{1}{8}}_{\{1,2\}},F^{\frac{1}{8}})\,,
	\end{equation}
	where $\hat{F}^{\frac{1}{8}}$ is the fused feature.

	Moving forward, we proceed with the $\frac{1}{4}$-scale aggregation stage. Instead of utilizing the feature $F^{\frac{1}{4}}$ directly, we opt for the $2\times$ upsampled counterpart $\hat{F}^{\frac{1}{8}\uparrow2}$ as the target-view feature. This choice is motivated by the presence of potential artifacts in the rendered view $I$. Given that $\hat{F}^{\frac{1}{8}}$ has already aggregated high-quality details from the reference views, it is deemed more appropriate for conducting correspondence estimation involving $F^{\frac{1}{4}}_i$ and the target view:
	%
	
	\begin{equation}
		\begin{aligned}
			& F^{\frac{1}{8}\uparrow2}_i = {\rm Bilinear}( \hat{F}^{\frac{1}{8}},2)\,, \\
			& M^{\frac{1}{4}}_{i} = [F^{\frac{1}{8}\uparrow2}_i,F^{\frac{1}{4}}_i,C_i^{\frac{1}{4}}]\,, \\
			& \hat{F}^{\frac{1}{4}}_i = \mathop{FDCN}(F^{\frac{1}{4}}_i,M^{\frac{1}{4}}_{i},C_i^{\frac{1}{4}})\,. \\
		\end{aligned}
		\label{eq:alignp2}
	\end{equation}
	Afterwards, we take another 3D convolution to mix the two aggregated features $\hat{F}^{\frac{1}{4}}_{\{1,2\}}$:
	
	\begin{equation}
		\hat{F}^{\frac{1}{4}} = \mathop{Conv3D}(\hat{F}^{\frac{1}{4}}_{\{1,2\}},F^{\frac{1}{4}}) \,.
	\end{equation}
	Finally, we conduct the third-level aggregation to obtain the fused feature $\hat{F}^{\frac{1}{2}}$, using similar processing steps as in the second stage.
	
	\vspace{0.03in}
	\noindent\textbf{Pyramid reconstruction and multi-scale supervision.}\label{sec:recons_p}
	We employ pyramid aggregated features $\hat{F}^{\{\frac{1}{8},\frac{1}{4},\frac{1}{2}\}}$ to generate multi-scale outputs. Initially, starting from $\hat{F}^{\frac{1}{8}}$, we employ convolutional blocks to obtain the lowest-scale output $\hat I^{\frac{1}{8}}$. Subsequently, we upsample this output and incorporate $\hat{F}^{\frac{1}{4}}$ to learn the image residue at a higher scale, yielding $\hat I^{\frac{1}{4}}$. By following this strategy, we ultimately predict the enhanced view $\hat I$. To improve reconstruction quality, we incorporate multi-scale supervision during training:
	
	\begin{equation}
		\begin{aligned}
			&{L_{\{{\frac{1}{8}},{\frac{1}{4}}\}}} =  || \hat{I}^{\{{\frac{1}{8}},{\frac{1}{4}}\}} - I_{gt}^{\{{\frac{1}{8}},{\frac{1}{4}}\}}||_1;\\
			&{L_f} =  || \hat{I} - I_{gt}||_1;\\
			&{L} = 0.1*{L_{\{0,1\}}} + {L_f},
		\end{aligned}
	\end{equation}
	where $I_{gt},I_{gt}^{\{\frac{1}{8},\frac{1}{4}\}}$ are the full-resolution and down-scaled ground truth views.
	
	\section{Experiment}
	\subsection{Implementation Details}
	Initially, we train the deep generative degradation simulator for 150K iterations. After this, we freeze the weights of both the deep generative degradation simulator and the optical flow model used in G-IVM for the next 300K iterations. Then we jointly train both the deep generative degradation simulator and G-IVM for additional 300K iterations, using a batch size of 16 and a cropped input size of $128\times 128$. We use the same data augmentation techniques as NeRFLiX~\cite{Zhou_2023_CVPR}, and employ an Adam optimizer and a Cosine annealing learning rate scheme.
	\subsection{Datasets and Metrics}
	Following NeRFLiX, we conduct experiments on three popular datasets: LLFF~\cite{mildenhall2019local}, Tanks and Temples~\cite{knapitsch2017tanks}, and Noisy LLFF Synthetic~\cite{mildenhall2020nerf}. The first two benchmarks have eight and five real-world scenes, respectively. Noisy LLFF Synthetic has eight virtual scenes, where we manually apply camera jetting to the precise camera poses to simulate the imperfect in-the-wild calibration.
	
	We evaluate our method using the PSNR~(${\color{red}\uparrow}$), SSIM~\cite{wang2004image}~(${\color{red}\uparrow}$) and LPIPS~\cite{zhang2018unreasonable}(${\color{red}\downarrow}$) metrics, consistent with the evaluation standards of NeRF models.

	\begin{figure*}[t]
		\begin{center}
			\includegraphics[width=\linewidth]{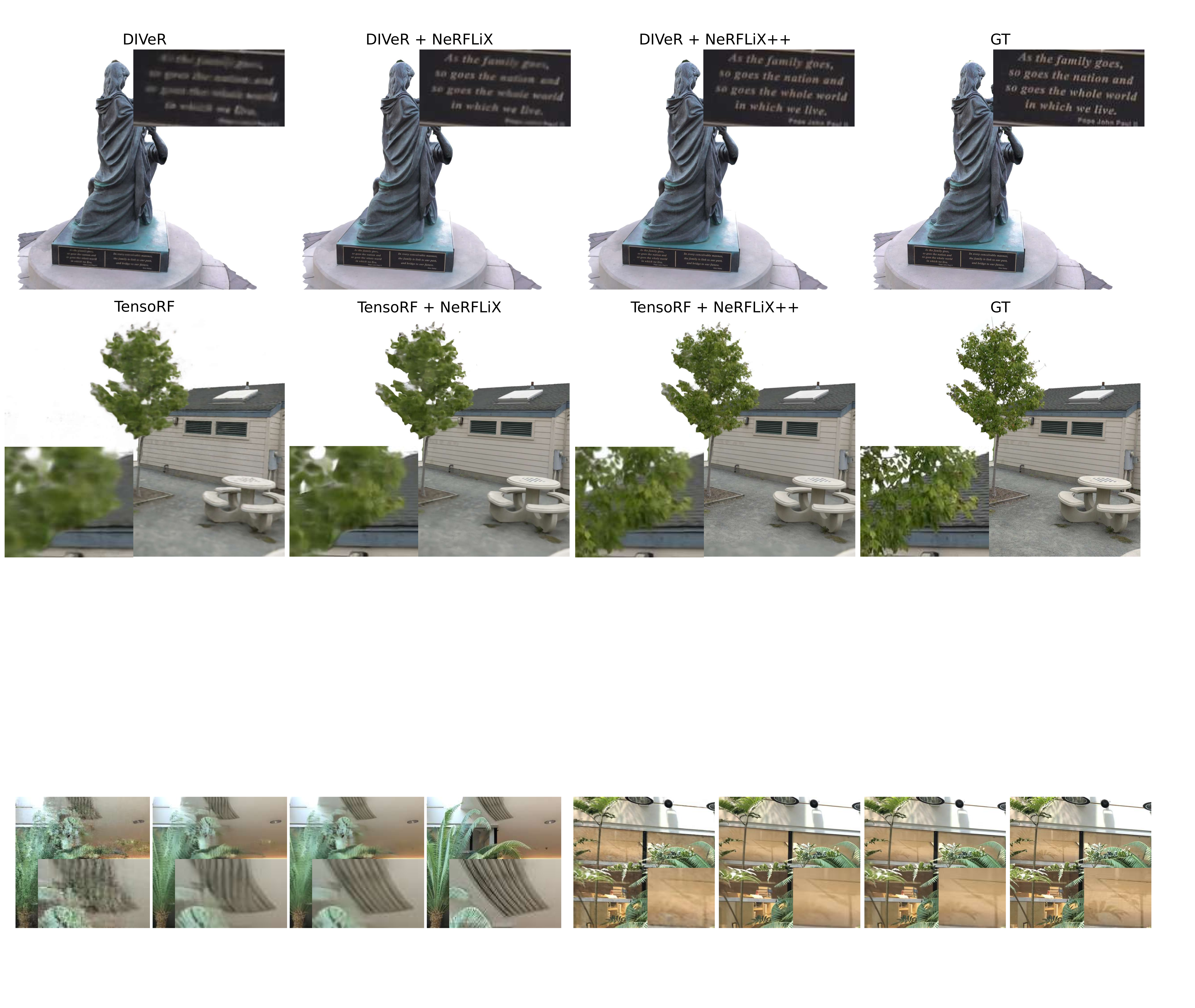} 
		\end{center}
		\vspace{-0.15in}
		\caption{Qualitative results of NeRFLiX and NeRFLiX++. It is observed that NeRFLiX++ is able to restore richer photo-realistic details than NeRFLiX, showing the superior performance of NeRFLiX++.
		}
		\label{fig:sotafig}
		\vspace{-0.1in}
	\end{figure*}
	\subsection{Improvements over SOTA NeRFs}
	We validate the effectiveness of NeRFLiX++ by consistently improving the performance of state-of-the-art NeRF models on diverse datasets. Furthermore, we conduct thorough quantitative and qualitative comparisons between NeRFLiX++ and NeRFLiX, while also assessing their respective inference efficiency.
	
	\vspace{0.05in}
	\noindent\textbf{LLFF.} In order to examine the enhancement potential of our NeRFLiX++, we investigate six representative models, including NeRF~\cite{mildenhall2020nerf}, TensoRF~\cite{tensorf}, Plenoxels~\cite{fridovich2022plenoxels}, NeRF-mm~\cite{wang2021nerf}, NLF~\cite{attal2022learning}, and RegNeRF~\cite{Niemeyer2021Regnerf}. Using rendered views~(as well as their reference views) of NeRF approaches as inputs to our model, we aim to further improve the synthesis quality. The quantitative results are provided in Table~\ref{tab:sota1}. Under both protocols, NeRFLiX++ exhibits comparable improvements compared to NeRFLiX, elevating the performance of NeRF models to unprecedented levels. For instance, NeRFLiX++ achieves significant improvements of \textit{0.61dB/0.025/0.054} in terms of PSNR/SSIM/LPIPS for the Plenoxels~\cite{fridovich2022plenoxels} dataset. Notably, NeRFLiX++ demonstrates \textit{2.4$\times$ smaller} model capacity and \textit{9.2$\times$ faster} processing speed for $1024\times1024$ images compared to NeRFLiX, as shown in Table~\ref{subtab:modelinference}. Furthermore, NeRFLiX++ achieves remarkable efficiency by processing ultra high-resolution frames of $2048\times2048$ in just 1.5 seconds.
	
	\begin{table}[t]
		\small
		\setlength{\tabcolsep}{3pt}
		\centering
		\begin{subtable}[b]{0.48\textwidth}
			\centering
			\begin{tabular}{l|c|c|c } 
				\hline
				Method                      & PSNR~(dB)${\color{red}\uparrow}$ &SSIM${\color{red}\uparrow}$ &LPIPS${\color{red}\downarrow}$   \\ \hline
				
				TensoRF~\cite{tensorf} {\footnotesize (ECCV'22)}              			 &  28.43      		     & 0.920  &0.142  \\ 
				TensoRF~\cite{tensorf} + NeRFLiX                 & {\color{blue}28.94} ($\color{red}\uparrow$ 0.51)         &   {\color{blue}0.930} &{\color{blue}0.120} \\ 
				TensoRF~\cite{tensorf} + NeRFLiX++                & {\color{red}29.24} ($\color{red}\uparrow$ 0.81)        &   {\color{red}0.937} &{\color{red}0.107} \\ \hline \hline

				DIVeR~\cite{wu2021diver}  {\footnotesize(CVPR'22)}             			 &  28.16      		     & 0.913  &0.145  \\ 
				DIVeR~\cite{wu2021diver} + NeRFLiX                 & {\color{blue}28.61} ($\color{red}\uparrow$ 0.45)         &   {\color{blue}0.924} &{\color{blue}0.127} \\ 
				DIVeR~\cite{wu2021diver} + NeRFLiX++                & {\color{red}28.85} ($\color{red}\uparrow$ 0.69)        &   {\color{red}0.933} &{\color{red}0.111} \\ \hline
			\end{tabular}
			
			\caption{Improvements over TensoRF and DIVeR on Tanks and Temples. Best and second best results are highlighted in {\color{red}red} and  {\color{blue}blue}.}
			\label{subtab:sotatanks}
			\vspace{0.05in}
		\end{subtable}
		\vspace{0.02in}
		\begin{subtable}[b]{0.48\textwidth}
			
			\centering
			\begin{tabular}{l|c|c|c } 
				\hline
				Method                      & PSNR~(dB)${\color{red}\uparrow}$ &SSIM${\color{red}\uparrow}$ &LPIPS${\color{red}\downarrow}$   \\ \hline
				
				TensoRF~\cite{tensorf} {\footnotesize (ECCV'22)}               			 &  22.83     		     & 0.881  &0.147  \\ 
				TensoRF~\cite{tensorf} + NeRFLiX                 &  {\color{blue}24.12} ($\color{red}\uparrow$ 1.29)        &   {\color{blue}0.913} &{\color{blue}0.092} \\ 
				TensoRF~\cite{tensorf} + NeRFLiX++                & {\color{red}25.39} ($\color{red}\uparrow$ 2.56)        &   {\color{red}0.926} &{\color{red}0.085} \\  \hline \hline

				\hline

				Plenoxels~\cite{fridovich2022plenoxels}  {\footnotesize(CVPR'22)}           			 &  23.69      		     & 0.882  &0.127  \\ 
				Plenoxels~\cite{fridovich2022plenoxels} + NeRFLiX                 &  {\color{blue}25.51} ($\color{red}\uparrow$ 1.82)        &   {\color{blue}0.920} &{\color{blue}0.084} \\ 
				TensoRF~\cite{tensorf} + NeRFLiX++                & {\color{red}26.82} ($\color{red}\uparrow$ 3.22)        &   {\color{red}0.930} &{\color{red}0.080} \\  \hline
			\end{tabular}
			\caption{Improvements over TensoRF and Plenoxels on noisy LLFF Synthetic.}
			\vspace{0.05in}
			\label{subtab:sotalego}
		\end{subtable}
		
		\begin{subtable}[b]{\columnwidth}
			
			\begin{tabular}{l|c } 
				\hline
				Method                      & PSNR (dB)${\color{red}\uparrow}$/SSIM${\color{red}\uparrow}$/LPIPS${\color{red}\downarrow}$  \\ \hline
				
				TensoRF~\cite{tensorf}(4 hours)             			 &26.73 / 0.839 / 0.204  \\ 
				TensoRF~\cite{tensorf}(\textbf{2 hours})             			 &26.18 / 0.819 / 0.230  \\ 
				\cite{tensorf}(\textbf{2 hours})  + NeRFLiX                 &{\color{blue}27.14} / {\color{blue}0.858} / {\color{red}0.165}  \\ 
				\cite{tensorf}(\textbf{2 hours})  + NeRFLiX++                 & {\color{red}27.15} / {\color{red}0.861} / {\color{blue}0.169} \\ \hline \hline

				\hline

				Plenoxels~\cite{fridovich2022plenoxels}(24 minutes)            			 &26.29 / 0.839 / 0.210  \\ 
				Plenoxels~\cite{fridovich2022plenoxels}(\textbf{10 minutes})                  &25.73 / 0.804 / 0.252\\
				
				\cite{fridovich2022plenoxels}(\textbf{10 minutes}) + NeRFLiX                 &{\color{red}26.60} / {\color{blue}0.847} / {\color{red}0.181} \\ 
				\cite{fridovich2022plenoxels}(\textbf{10 minutes}) + NeRFLiX++                & {\color{blue}26.57} / {\color{red}0.849} / {\color{red}0.181} \\ \hline

			\end{tabular}
			\caption{Improvements over TensoRF and Plenoxels trained with half of the 
				recommended iterations on LLFF~\cite{mildenhall2019local} under LLFF-P1.}
			\label{subtab:sotahalf}
		\end{subtable}
		\caption{Quantitative evaluation of improvements of NeRFLiX and NeRFLiX++ for various NeRFs.}\label{tab:sotasynthetic}
		\vspace{-0.15in}
	\end{table}

	\vspace{0.05in}
	\noindent\textbf{Tanks and Temples.} 
	Compared with the LLFF, it has large variations of camera viewpoints. As a result, even recent advanced NeRF models, \textit{e.g.}, TensoRF~\cite{tensorf} and DIVeR~\cite{wu2021diver},  fail to synthesize high-quality results.
	As depicted in Table~\ref{subtab:sotatanks}, both NeRFLiX and NeRFLiX++ demonstrate substantial performance improvements across these models. Particularly, NeRFLiX++ exhibits enhanced generalization capabilities, resulting in more significant performance gains. For example, NeRFLiX++ achieves notable improvements of \textit{0.81dB}/\textit{0.017}/\textit{0.035} on PSNR/SSIM/LPIPS for the TensoRF~\cite{tensorf} model.

	\vspace{0.05in}
	\noindent\textbf{Noisy LLFF Synthetic.} Apart from in-the-wild benchmarks above, we also demonstrate the enhancement capability of our model on noisy LLFF Synthetic. From the results shown in Table~\ref{subtab:sotalego}, we see that our NeRFLiX++ yields substantial improvements upon two SOTA NeRF models.
	
	\vspace{0.05in}
	\noindent {\color{black} 
		{\textbf{Performance v.s. computation cost Trade-off.} We investigate the trade-off between performance enhancement and computational overhead when using NeRFLiX and NeRFLiX++. Fig.~\ref{fig:speedcom} illustrates the relationship between PSNR, inference time, and memory usage. NeRFLiX++ notably improves upon state-of-the-art NeRF models while maintaining acceptable processing speed for high-resolution inputs. For instance, NeRFLiX++ enhances the NeRF model by \textbf{0.75dB} with only a \textbf{$0.77\%$} increase in inference time (0.43s out of 56.04s for processing $1024\times1024$ frames).}
	}

	\vspace{0.05in}
	\noindent\textbf{Qualitative results.} Fig.~\ref{fig:sotafig} presents qualitative examples for visual assessment. The results demonstrate that NeRFLiX++ effectively restores clearer image details while significantly reducing NeRF-style artifacts in the rendered images, highlighting the efficacy of our approach.
	
	\vspace{-0.05in}
	\subsection{Training Acceleration for NeRF Models}
	In this section, we show how NeRFLiX(++) makes it possible for NeRF models to produce better results even with a 50$\%$ reduction in training time. To be more precise, we make use of NeRFLiX and NeRFLiX++ to improve the rendered images of two SOTA NeRF models after training them with half the training period specified in the publications. The enhanced results \textit{outperform} the counterparts with full-time training, as shown in Table~\ref{subtab:sotahalf}.  Notably, both NeRFLiX and NeRFLiX++ have reduced the training period for Plenoxels~\cite{fridovich2022plenoxels} from 24 minutes to 10 minutes while also consistently improving the quality of the rendered images.
	
	\vspace{-0.05in}
	\subsection{Ablation Study}
	In this section, we conduct comprehensive experiments on LLFF~\cite{mildenhall2019local} under the LLFF-P1 protocol to analyze each of our designs. We use TensoRF~\cite{tensorf} as our baseline\footnote{The TensoRF results~(26.70dB/0.838/0.204) that we examined slightly differ from the published results~(26.73dB/0.839/0.204).}.
	
	\begin{figure}[ht]
		\centering
		\includegraphics[width=\columnwidth]{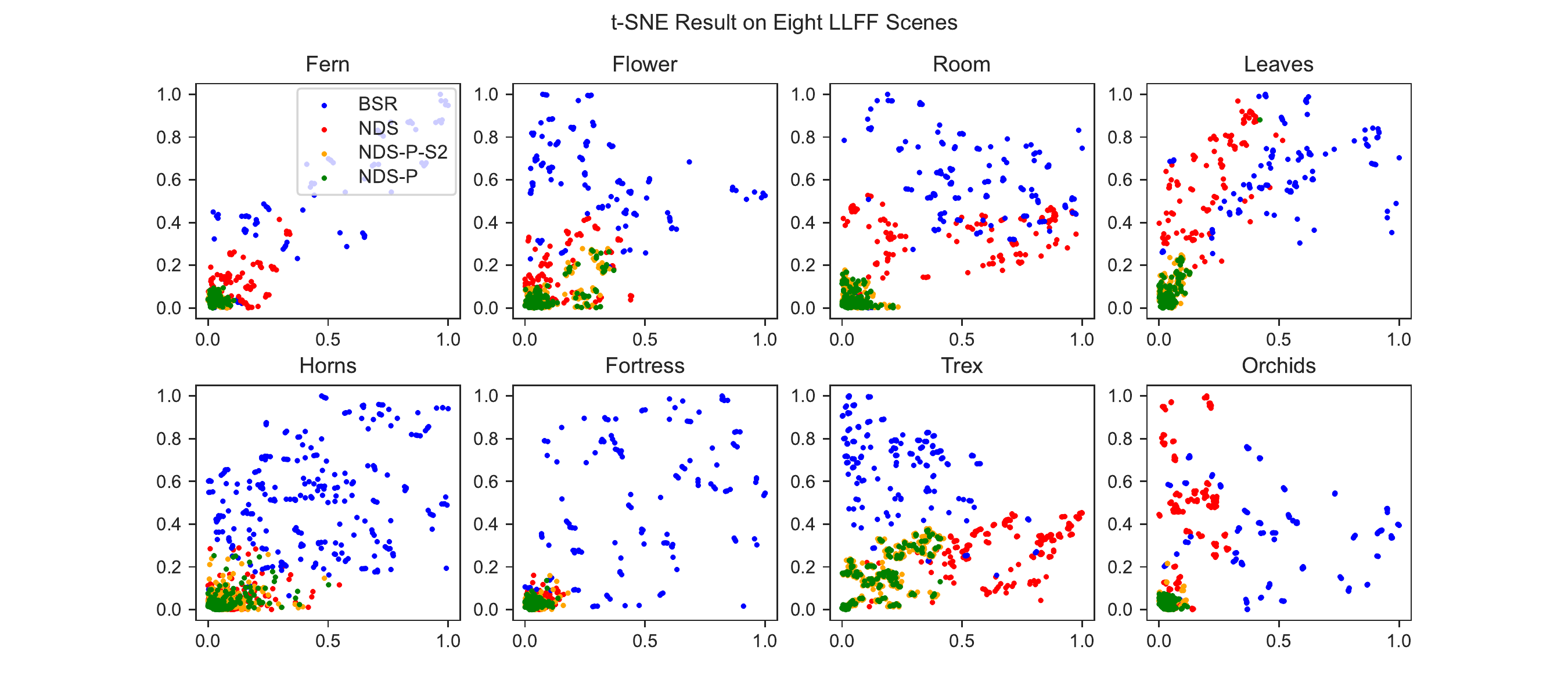} 
		
		\caption{\color{black}{The normalized differences between simulated images from the four degradation models and real NeRF-rendered images. Smaller values signify closer distribution similarity. ``BSR", ``NDS", ``NDS-P-S2", and ``NDS-P" denote BSR~\cite{zhang2021designing}, NeRFLiX, NeRFLiX++ with only deep generative degradation, and NeRFLiX++, respectively.}
		} 
		
		\label{fig:tsne}
	\end{figure} %
	\vspace{0.03in}
	\noindent\textbf{Data simulation quality.}
	{\color{black}
		To assess simulation quality, we measure the distribution similarity between real-rendered frames and simulated frames generated by various degradation models: BSR~\cite{zhang2021designing}, NeRFLiX, deep generative degradation in NeRFLiX++, and NeRFLiX++. We employ t-SNE~\cite{van2008visualizing} to visualize deep image features extracted using Inception-v3~\cite{szegedy2016rethinking}, as shown in Fig.~\ref{fig:tsne}. Notably, our two-stage simulator, NeRFLiX++, produces simulated data statistically closest to real rendered images, surpassing other models. Quantitative analysis further supports this finding in Table~\ref{subtab:dev2} and Table~\ref{table:nds-p-c}.
		
		In addition, we perform a thorough analysis to assess the individual components of the degradations, namely the human-crafted simulator (H-C.S) and the deep generative simulator (D-G.S). We train eight models, gradually incorporating more degradation strategies. The results in Table~\ref{table:nds-p-c} highlight the importance of each degradation component in achieving desirable results. Notably, the model utilizing the generative degradation simulator outperforms the one relying solely on the manual degradation simulator.
	}
	
	
	\begin{table}[t]
		\small
		\centering
		\setlength{\tabcolsep}{2pt}
		\begin{tabular}{c|c|c|c|c|c|c } 
			\hline
			Metrics &Baseline &D.Models   & SwinIR   &DATSR              & EDVR   &VST  \\ \hline
			\multirow {3}{*}{PSNR} &\multirow {3}{*}{26.70} &\cellcolor{blue!15}BSR        &\cellcolor{blue!15}26.20$\downarrow$   &\cellcolor{blue!15}25.99$\downarrow$               	&\cellcolor{blue!15}26.01$\downarrow$  &\cellcolor{blue!15}25.19$\downarrow$ \\
			& &\cellcolor{green!15}NDS         &\cellcolor{green!15}26.82$\color{red}\uparrow$   &\cellcolor{green!15}26.84$\color{red}\uparrow$               	&\cellcolor{green!15}26.88$\color{red}\uparrow$  &\cellcolor{green!15}26.79$\color{red}\uparrow$ \\
			& &\cellcolor{yellow!15}NDS-P        &\cellcolor{yellow!15}26.85$\color{red}\uparrow$  &\cellcolor{yellow!15}26.90$\color{red}\uparrow$              	&\cellcolor{yellow!15}26.98$\color{red}\uparrow$  &\cellcolor{yellow!15}26.94$\color{red}\uparrow$ \\ \hline
			
			\multirow {3}{*}{SSIM} &\multirow {3}{*}{0.838}  &\cellcolor{blue!15}BSR        &\cellcolor{blue!15}0.834$\downarrow$   &\cellcolor{blue!15}0.826$\downarrow$               	&\cellcolor{blue!15}0.819$\downarrow$  &\cellcolor{blue!15}0.705$\downarrow$ \\
			& &\cellcolor{green!15}NDS        &\cellcolor{green!15}0.845$\color{red}\uparrow$ &\cellcolor{green!15}0.843$\color{red}\uparrow$                &\cellcolor{green!15}0.847$\color{red}\uparrow$   &\cellcolor{green!15}0.842$\color{red}\uparrow$ \\
			& &\cellcolor{yellow!15}NDS-P       &\cellcolor{yellow!15}0.847$\color{red}\uparrow$ &\cellcolor{yellow!15}0.847$\color{red}\uparrow$                &\cellcolor{yellow!15}0.850$\color{red}\uparrow$   &\cellcolor{yellow!15}0.849$\color{red}\uparrow$ \\
			\hline 
			
		\end{tabular}
		\vspace{-0.05in}
		\caption{Quantitative results of utilizing different degradations in the existing image and video processing models including SwinIR~\cite{liang2021swinir}, DATSR~\cite{cao2022datsr}, EDVR~\cite{wang2019edvr} and VSR~\cite{liu2022video}. We re-train these four models on three simulated datasets produced by BSR~\cite{zhang2021designing}, NDS~\cite{Zhou_2023_CVPR} and our proposed NDS-P. $\color{red}\uparrow$/$\downarrow$ indicate the model achieves better/worse performance compared with baseline~(TensoRF).}
		\label{subtab:dev2}
		
		%
		%
		%
		\vspace{-0.1in}
	\end{table}
	
	\begin{table}[h]
		\footnotesize
		\setlength{\tabcolsep}{1.5pt}
		\begin{center}
			
			\begin{tabular}{c|c|c|c|c|c|c|c|c|c } 
				\hline
				Types     &SGN &Re-pos. & A-Blur  &I-J &L-C  &RA &S2  &Metrics$\uparrow$ &t-SNE Diff.$\downarrow$\\ \hline
				
				\multirow{6}{*}{H-C.S}    		&\cellcolor{gray!15}\cmark	 &\cellcolor{gray!15}         & \cellcolor{gray!15}       &\cellcolor{gray!15}       &\cellcolor{gray!15}       & \cellcolor{gray!15}      &\cellcolor{gray!15}      &\cellcolor{gray!15}26.74/0.845&\cellcolor{gray!15}0.496  \\ 
				&\cmark	 &\cmark   &        &       &       &       &      &26.81/0.849&0.485  \\
				&\cellcolor{gray!15}\cmark	 &\cellcolor{gray!15}\cmark   &\cellcolor{gray!15}\cmark  &\cellcolor{gray!15}       &\cellcolor{gray!15}       &\cellcolor{gray!15}       &\cellcolor{gray!15}      &\cellcolor{gray!15}26.87/0.853&\cellcolor{gray!15}0.456  \\ 
				&\cmark	 &\cmark   &\cmark  &\cmark &       &       &	   &26.95/0.854	&0.413		 \\
				&\cellcolor{gray!15}\cmark	 &\cellcolor{gray!15}\cmark   &\cellcolor{gray!15}\cmark  &\cellcolor{gray!15}\cmark &\cellcolor{gray!15}\cmark &\cellcolor{gray!15}       &\cellcolor{gray!15}	   &\cellcolor{gray!15}27.09/0.856 &\cellcolor{gray!15}0.358            \\
				&\cmark	 &\cmark   &\cmark  &\cmark &\cmark &\cmark  & & 27.14/0.858 &0.331                   \\ \hline
				D-G.S               	&	     &         &        &       &       &       &\cmark   &27.20/0.860 &0.289           \\ \hline
				All         	&\cmark	 &\cmark   &\cmark  &\cmark &\cmark &\cmark &\cmark    &\textbf{27.38}/\textbf{0.866}  &\textbf{0.264}       \\  
				
				\hline
				
				\hline
			\end{tabular}
		\end{center}
		\caption{\color{black} PSNR/SSIM results and simulation quality for various degradations in our two-stage degradation simulator. ``H-C.S" and ``D-G.S" represent the human-crafted simulator and deep generative simulator, respectively. ``All" refers to our complete configuration, which incorporates both the manual and deep degradation simulator. ``SGN", ``Re-pos.", ``A-Blur", ``I-J", ``L-C" and ``RA" denote Splatted Gaussian noise, Re-positioning,  Anisotropic blur, illumination jetting, lightness compression schemes, and region adaptive strategy, while ``S2" refers to the deep generative degradation simulator.}
		
		\label{table:nds-p-c}
		\vspace{-0.1in}
	\end{table}
	
	\begin{table}[t]
		\setlength{\tabcolsep}{4pt}
		\begin{center}
			\begin{tabular}{l|c||c|c|c|c|c } 
				\hline
				Supervision	      & L1	       &$K=1$        &$K=3$       &$K=5$     &$K=10$  &$K=20$        \\ \hline
				PSNR (dB) 	&27.03         &27.21        &27.35         &27.38    &27.38       &27.37              \\ \hline
				SSIM		&0.851         &0.859        &0.865         &0.866    &0.866       &0.866 \\ \hline
				
			\end{tabular}
		\end{center}
		\vspace{-0.1in}
		\caption{Impact of different similarity patch numbers~($K$) for WKS. Moreover, we also include another model trained with L1 loss to validate the effectiveness of our WKS supervision. }
		\label{table:topk}
	\end{table}

	\vspace{0.03in}
	\noindent\textbf{Weighted top-$K$ similarity loss~(WKS).}  
	We evaluate the performance of our proposed WKS. For comparison, we train an additional G-IVM model using the conventional L1 loss as supervision. The results in Table~\ref{table:topk} demonstrate that this model achieves significantly inferior performance compared to the models trained with our proposed WKS. This outcome emphasizes the effectiveness of WKS for deep degradation training. Furthermore, we investigate the influence of different numbers (K) of similar patches in WKS supervision. We train four additional G-IVM models. As indicated in Table~\ref{table:topk}, we observe progressive improvements in PSNR values as the number of similar patches increases from $K=1$ to $K=5$, after which the improvements saturate. This behavior is expected since image patches with relatively small similarities contribute less to the overall performance.
	\vspace{0.03in}
	{\color{black}
		\noindent\textbf{Pyramid fusion in G-IVM.}  To leverage multi-scale contextual information from inter-viewpoint frames, we introduce a pyramid-guided aggregation structure. Table~\ref{table:py} illustrates that the incorporation of additional aggregation and reconstruction levels consistently enhances the final performance. Notably, our full model (Model-A) achieves the highest PSNR/SSIM scores.
	}
	
	\vspace{0.03in}
	\noindent\textbf{Flow guidance in G-IVM.} 
	To address distinct viewpoint changes in high-resolution frames, we introduce the utilization of coarse optical flow for guiding the aggregation process. In order to assess the significance of this strategy, we train an additional model referred to as ``NG-IVM" under the same experimental setup, but without utilizing optical flow guidance. The results presented in Table~\ref{table:flow} clearly indicate that our guided inter-viewpoint mixer outperforms the NG-IVM model by a substantial margin, highlighting the effectiveness of our design.
	

	\begin{table}[t]
		
		\setlength{\tabcolsep}{2pt}
		
		\begin{subtable}[b]{0.46\textwidth}
			\centering
			\begin{tabular}{c|c|c|c|c|c|c|c|c } 
				\hline
				Models    &A$_3$ &R$_3$ &A$_2$ &R$_2$ &A$_1$ &R$_1$   &Metrics&Param./Time/Mem. \\ \hline
				
				\rowcolor{gray!15}	A             		&\cmark	 &\cmark       		     &\cmark &\cmark      & \cmark  &\cmark  &\textbf{27.38}/\textbf{0.866} &{\color{black}14.4M/433ms/7.1GB}  \\ 
				B             		&\cmark	 &\cmark       		     &\cmark &\cmark      &\cmark   &  &27.26/0.863 &{\color{black}12.3M/429ms/7.1GB}  \\ 
				\rowcolor{gray!15}	C             		&\cmark	 &\cmark       		     &\cmark &\cmark      &   &  &27.11/0.856 &{\color{black}10.8M/412ms/7.1GB}  \\ 
				D             		&\cmark	 &\cmark       		     &\cmark &      &   &  &27.03/0.854 &{\color{black}9.7M/408ms/7.0GB}  \\ 
				\rowcolor{gray!15}	E                 	&\cmark	 &\cmark       		     & &      &   &  &26.90/0.849  &{\color{black}8.1M/367ms/6.9GB}\\ 
				F            		&\cmark	 &     &     &      &   &  &26.84/0.848 &{\color{black}7.1M/323ms/6.9GB}\\ 
				
				\hline
				
				\hline
			\end{tabular}
			\caption{\color{black}
				{Comparison of aggregation and reconstruction strategies, A$_{\{1,2,3\}}$ and R$_{\{1,2,3\}}$, at different levels. Inference cost is calculated for a $1024\times 1024$ resolution on an NVIDIA RTX 3090. }
			}
			\label{table:py}
			\vspace{0.1in}
		\end{subtable}
		
		\begin{subtable}[b]{0.22\textwidth}
			\begin{tabular}{l|c|c } 
				\hline
				Setting		&PSNR&SSIM       \\ \hline
				w/o Flow 		            &27.21&0.860            \\ 
				w/ Flow 		&\textbf{27.38}&\textbf{0.866}              \\ \hline
			\end{tabular}
			\caption{Impact of flow guidance.}
			\label{table:flow}
		\end{subtable}
		\quad \hfill
		\begin{subtable}[b]{0.22\textwidth}
			\begin{tabular}{l|c|c } 
				\hline
				Target		&PSNR&SSIM       \\ \hline
				$F^l$ 		            &26.97&0.853            \\ 
				$\hat{F}^{l-1}$ 		&\textbf{27.38}&\textbf{0.866}                \\ \hline
			\end{tabular}
			\caption{Impact of aligning targets.}
			\label{table:reff}
		\end{subtable}
		
		\caption{Experimental analyses to understand the roles of pyramid fusion, flow guidance and different aligning targets and present the quantitative results.}
		\vspace{-0.1in}
		\label{table:scale}
	\end{table}
	
	\begin{table}[h]
		\setlength{\tabcolsep}{4pt}
		\begin{center}
			\begin{tabular}{l|c|c|c|c|c } 
				\hline
				Scales		&$\times 1$        &$\times 2$      &$\times 4$     &$\times 8$ &$\times 16$        \\ \hline
				PSNR (dB) 		&27.25         &27.28  &\textbf{27.38}         &27.21 &27.17                      \\ \hline
				SSIM		&0.863         &0.863         &\textbf{0.866}  &0.862 &0.860          \\ \hline
				{\color{black}Speed (ms)}  	&790         &453         &433  &412 &401 \\ 
				{\color{black}Memory (GB)}     &7.4   &7.1         &7.1  &7.1 &7.1 \\ \hline
			\end{tabular}
		\end{center}
		\vspace{-0.1in}
		\caption{\color{black} Impacts of performing coarse correspondence estimation at different image sizes. Inference cost is calculated for a $1024\times 1024$ resolution on an NVIDIA RTX 3090.}
		\label{table:ablationall}
			\vspace{-0.05in}
	\end{table}
	
	\begin{figure*}[t]
		\begin{center}
			\includegraphics[width=\linewidth]{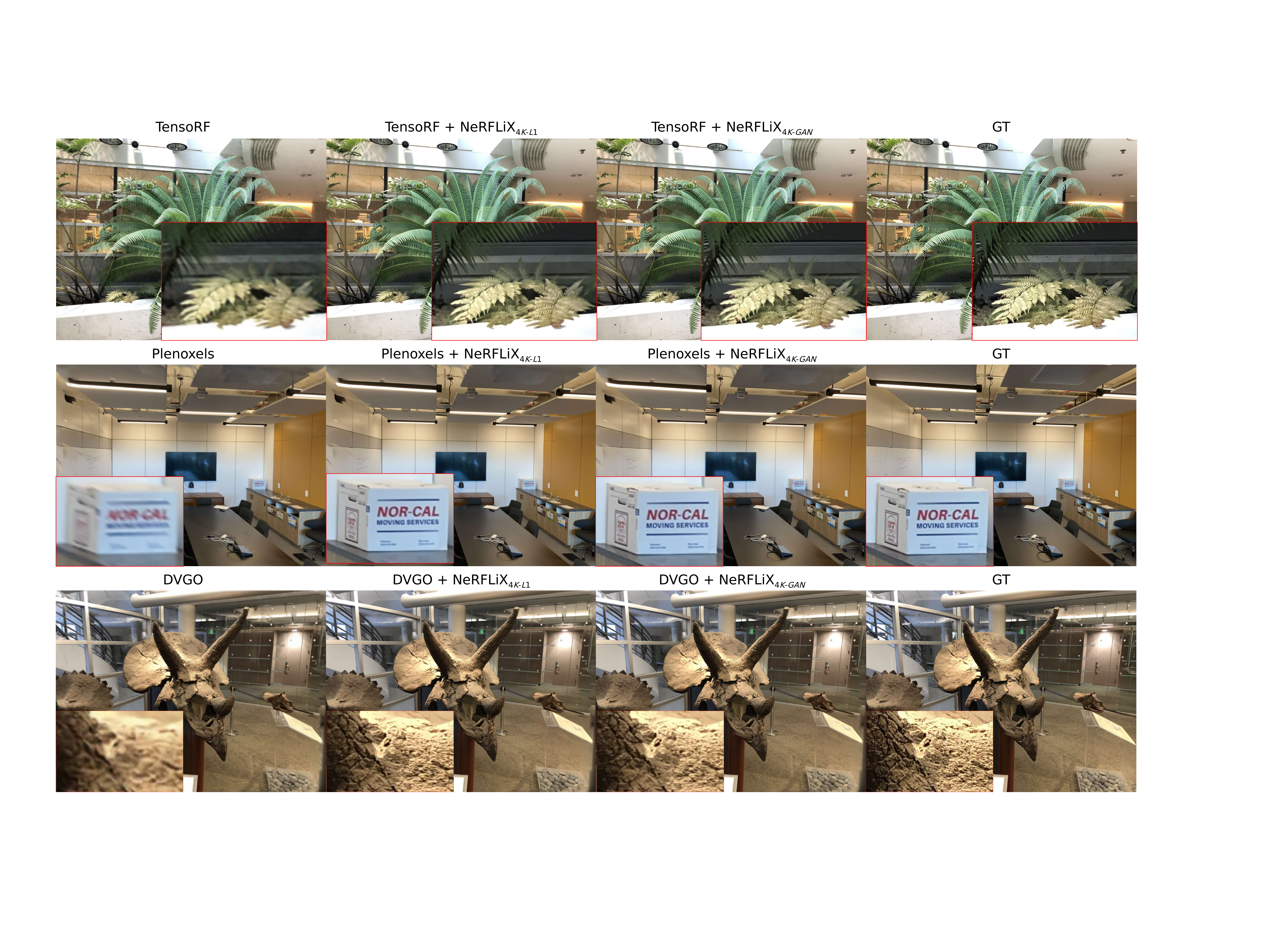} 
		\end{center}
		\caption{Qualitative results of restoring 4K images from noisy 1K frames produced by TensoRF, Plenoxels, DVGO. It is clear that NeRFLix++$_{4K}$ is capable of fusing high-quality reference views to generate natural image textures.
		}
		\label{fig:nerf4k}
	\end{figure*}
	
	\vspace{0.03in}
	\noindent\textbf{Aligning targets in G-IVM.} 
	In the $l$-th level fusion, we deviate from existing approaches~\cite{wang2019edvr,zhou2022revisiting,li2020mucan,chan2021basicvsr++} that treat $F^l$ as the target feature. Instead, apart from the first-level alignment~($l=0$), we propose using the previously aggregated feature $\hat{F}^{l-1}(l>0)$. To validate the effectiveness of this design choice, we compare these two strategies, and the results are presented in Table~\ref{table:reff}. Our fusion strategy outperforms the existing solution in terms of PSNR, SSIM, and LPIPS, indicating that our design is better suited for NeRF-agnostic restoration tasks.
	
	\vspace{0.03in}
	\noindent\textbf{Correspondence estimation size.} 
	{\color{black}
		{In Section~\ref{sec:introduction}, we discussed the potential advantages of employing coarse correspondence estimation at a reduced resolution (scaled down by a factor of $\times 4$). Here, we present comparisons at other scales ($\times 1, \times 2, \times 8, \times 16$). Table~\ref{table:scale} provides results for four NeRFLiX++ models (NeRFLiX++$_{\{\times 1, \times 2, \times 8, \times 16\}}$) in addition to the default setting (referred to as "NeRFLiX++$_{\times 4}$"). Notably, NeRFLiX++$_{\times 4}$ achieves superior results in terms of PSNR and SSIM. It is worth noting that larger downscaling ratios (\textit{e.g.,} $\times 16$) may compromise accurate guidance performance for high-resolution aggregations.
			
			Additionally, we assess the computational cost associated with different alignment scales. It is observed that conducting high-resolution optical flow estimation adds an extra inference time of approximately 350ms. Our $\times$4 model achieves a favorable balance between performance and efficiency when compared to other models.}
	}

	\begin{figure*}[h]
		\centering
		\includegraphics[width=\linewidth]{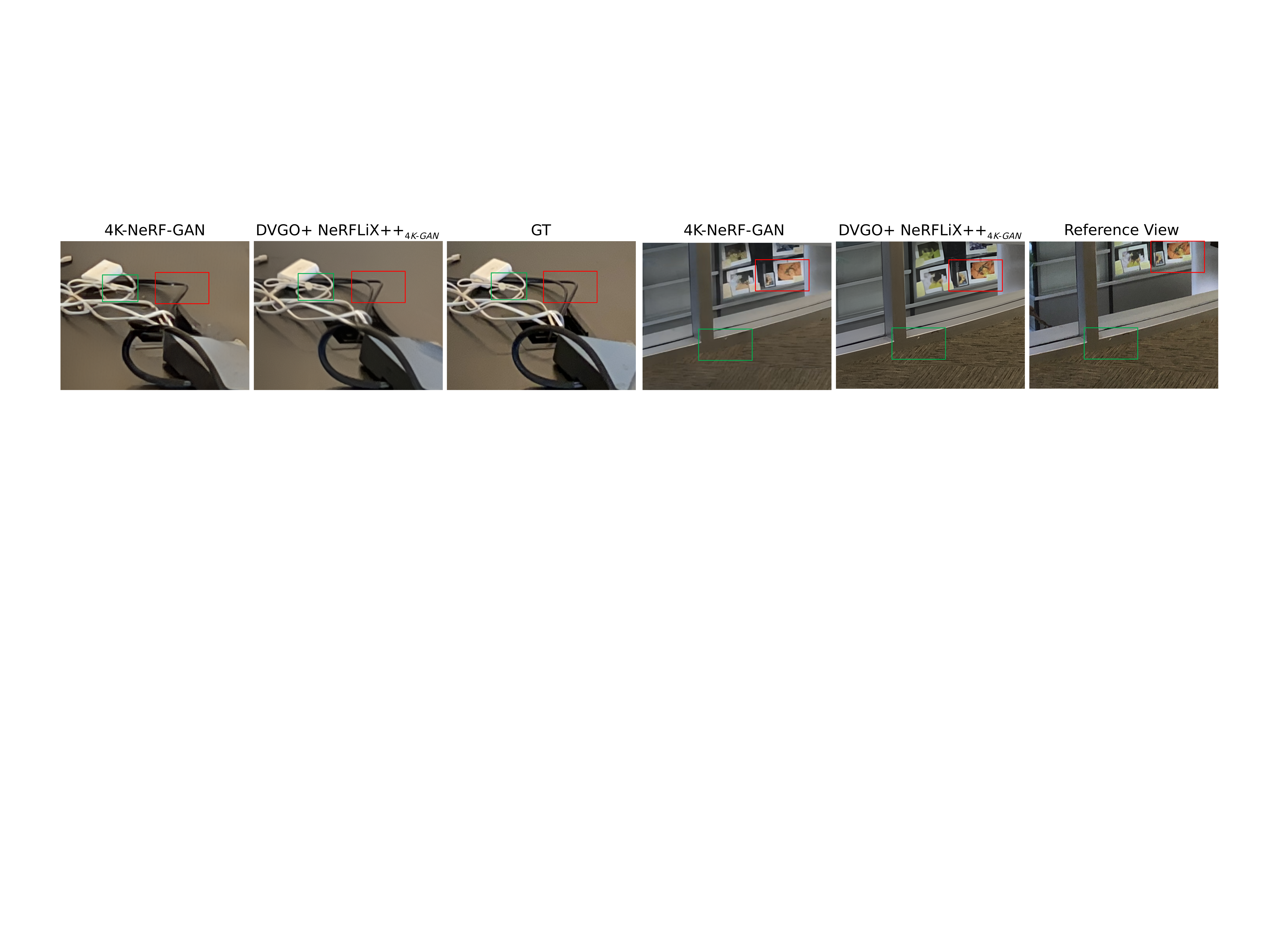} 
		\vspace{-0.2in}
		\caption{ Visual comparisons of 4K-NeRF and our NeRFLiX++$_{4K-GAN}$. While 4K-NeRF-GAN generates incorrect structures, our approach produces photo-realistic image details that closely resemble the ground truth or reference views (for novel viewpoints).
		} 
		\vspace{-0.05in}
		\label{fig:nerf-4k}
	\end{figure*} %
	
	\section{NeRFLiX++ for 4K images}
	
	In addition to the common challenges encountered in low-resolution novel view rendering, such as artifacts and blurriness, rendering high-resolution images, \textit{i.e.}, 4K resolution, using existing NeRF models poses significant computational resource requirements. Even with highly optimized data structures like tensor decomposition employed in TensoRF~\cite{tensorf}, training TensoRF models for 2K and 4K images on an NVIDIA RTX 3090 remains impractical due to limitations in GPU memory.

	In this section, we investigate the potential of utilizing NeRFLiX++ to super-resolve and enhance low-resolution images generated by different NeRF models, thereby producing high-quality 4K results. We first define the problem and then discuss the modifications made to the G-IVM model. Subsequently, we perform quantitative and qualitative analyses to assess the effectiveness of NeRFLiX++ for 4K images.
	
	%
	
	\subsection{Problem Formulation} 
	Given a low-resolution (1K) target frame $I$ generated by NeRF models and its two 4K reference views $\{I_1, I_2\}$, NeRFLiX++$_{4K}$ aims to restore a 4K output with photo-realistic details.

	\begin{table}[t]
		\setlength{\tabcolsep}{3pt}
		\begin{center}
			\begin{tabular}{c|c|c|c|c|c|c|c|c } 
				\hline
				Baseline                      &M1  &M2   &M3   &M4 &M5 &PSNR$\uparrow$   &SSIM$\uparrow$ &LPIPS$\downarrow$\\ \hline
				TensoRF             		&\cmark	&     &     &    &   		     &25.09   &0.873&0.43  \\
				TensoRF             		&	 	&\cmark     &     &    &   		     &25.54   &0.886&0.42  \\
				TensoRF             		&	 	&     &\cmark     &    &   		     &25.54   &0.885&0.40  \\
				TensoRF             		&	 	&     &     &\cmark    &   		     &\textbf{25.89}   &\textbf{0.896}&0.35  \\
				TensoRF             		&	 	&     &     &    &\cmark   		     &25.45   &0.885&\textbf{0.18}  \\  
				\hline \hline
				Baseline                       &M1  &M2   &M3   &M4 &M5 &PSNR$\uparrow$   &SSIM$\uparrow$ &LPIPS$\downarrow$\\ \hline
				Plenoxels             		&\cmark	&     &     &    &   		     &24.81   &0.873&0.47  \\
				Plenoxels             		&	 	&\cmark     &     &    &   		     &25.22   &0.885&0.42  \\
				Plenoxels             		&	 	&     &\cmark     &    &   		     &25.25   &0.884&0.42  \\
				Plenoxels             		&	 	&     &     &\cmark    &   		     &\textbf{25.50}   &\textbf{0.895}&0.35  \\
				Plenoxels             		&	 	&     &     &    &\cmark   		     &25.09   &0.884&\textbf{0.18}  \\  
				\hline \hline
				Baseline                      &M1  &M2   &M3   &M4 &M5 &PSNR$\uparrow$   &SSIM$\uparrow$ &LPIPS$\downarrow$\\ \hline
				DVGO             		&\cmark	&     &     &    &   		     &25.20   &0.869&0.50  \\
				DVGO             		&	 	&\cmark     &     &    &   		     &25.86   &0.888&0.43  \\
				DVGO             		&	 	&     &\cmark     &    &   		     &25.86   &0.888&0.42  \\
				DVGO             		&	 	&     &     &\cmark    &   		     &\textbf{26.21}   &\textbf{0.900}&0.35  \\
				DVGO             		&	 	&     &     &    &\cmark   		     &25.84   &0.889&\textbf{0.19}  \\  
				\hline \hline
				
			\end{tabular}
		\end{center}
		\vspace{-0.1in}
		\caption{\color{black}
			{Quantitative performance of various up-scaling strategies and image/video restoration methods, including M1 (bicubic interpolation, denoted as ``Bi"), M2 (NeRFLiX $\circ$ Bi), M3 (NeRFLiX++ $\circ$ Bi), M4 (NeRFLiX++$_{4K-L1}$, supervised by L1 loss), and M5 (NeRFLiX++$_{4K-GAN}$ with an adversarial loss). The symbol $\circ$ indicates that the two methods were executed sequentially. We also include three baseline methods: TensoRF~\cite{tensorf}, Plenoxels~\cite{fridovich2022plenoxels}, and DVGO~\cite{Sun_2022_CVPR}.}
		}
		\vspace{-0.1in}
		\label{table:nerf4k}
	\end{table}
	\subsection{Framework} 
	The framework of our proposed G-IVM is depicted in Fig.~\ref{fig:GIVMF}. To enable the restoration of 4K images, we make minimal modifications to the G-IVM framework. 
	
	\vspace{0.03in}
	\noindent\textbf{Encoder.} 
	To accommodate the resolution difference between the input frame $I \in \mathcal{R}^{H \times W}$ and its reference views $\{I_1, I_2\} \in \mathcal{R}^{4H \times 4W}$, we introduce the following adjustments. For encoder-1, we incorporate two convolutional layers with a stride of 2, resulting in down-sized reference features $F_{\{1,2\}}^{\{\frac{1}{4},\frac{1}{2}, 1\}} \in \mathcal{R}^{\{H \times W, 2H \times 2W, 4H \times 4W\}}$. Encoder-2 does not involve any down-sampling. Consequently, the two lowest-resolution reference features $F_{\{1,2\}}^{\frac{1}{4}} \in \mathcal{R}^{H \times W}$ match the spatial resolution of the input feature $F \in \mathcal{R}^{H \times W}$.

	We train two NeRFLiX++$_{4K}$ models, one using L1 loss (NeRFLiX++$_{4K-L1}$) and the other using a combination of L1 and GAN losses (NeRFLiX++$_{4K-GAN}$).
	%
	
	\vspace{0.03in}
	\noindent\textbf{Implementation details.} Compared to the original NeRFLiX++, when utilizing a 1K rendered frame as input, we substitute the two 1K reference frames with their 4K counterparts (obtained from LLFF-T\footnote{LLFF-T provides images at a 4K resolution}), while keeping other training details unchanged.
	
	For samples obtained from the Vimeo dataset, we initially down-sample the input frame by a factor of $\times$4, thereby establishing the same setup as LLFF-T. In other words, this configuration involves a low-resolution input view and two high-resolution reference views.
	%
	
	\subsection{Improvements over NeRFs for 4K Images}
	
	To evaluate the effectiveness of NeRFLiX++$_{4K}$, we conduct experiments using different restoration methods (M1-M5) to generate 4K images from low-resolution inputs produced by three state-of-the-art NeRF models: TensoRF~\cite{tensorf}, Plenoxels~\cite{fridovich2022plenoxels}, and DVGO~\cite{Sun_2022_CVPR}. The results in Table~\ref{table:nerf4k} demonstrate that all models involving NeRFLiX and NeRFLiX++ (M2-M5) outperform simple bicubic up-sampling (M1), indicating the restoration capability of NeRFLiX and NeRFLiX++. Particularly, NeRFLiX++$_{4K-L1}$ (M4) and NeRFLiX++$_{4K-GAN}$ (M5) achieve the best performance in terms of PSNR, SSIM, and LPIPS. Furthermore, Fig.~\ref{fig:nerf4k} visually demonstrates that NeRFLiX++$_{4K-L1}$ (M4) produces high-quality 4K frames with clearer textures and reduced rendering artifacts. Meanwhile, NeRFLiX++$_{4K-GAN}$ (M5) generates more high-frequency details and sharper edges, resulting in visually appealing results.
	
	\begin{table}[t]
		\small

		\begin{subtable}[b]{0.46\textwidth}
			\centering
			\footnotesize
			\setlength{\tabcolsep}{1pt}
			\begin{tabular}{l|c|c||c|c} 
				\hline
				Method &4K-NeRF$_{L1}$  &NeRFLiX++$_{4K-L1}$ &4K-NeRF$_{GAN}$ &NeRFLiX++$_{4K-GAN}$ \\ \hline
				PSNR$\uparrow$  &25.44 &\textbf{26.21} &24.71 &25.84\\
				SSIM$\uparrow$ &0.883 &\textbf{0.900} &0.871 &0.889 \\
				LPIPS$\downarrow$ &0.41 &0.35 &0.24 &\textbf{0.19} \\ \hline
			\end{tabular}
			\caption{Quantitative comparisons between 4K-NeRF~\cite{wang20224k} and our method on 4K NeRF-rendered images.}
			\label{subtab:4k-nerf}
			\vspace{0.1in}
		\end{subtable}
		\hfill
		\begin{subtable}[b]{0.46\textwidth}
			\centering
			\setlength{\tabcolsep}{2pt}
			\begin{tabular}{l|c|c} 
				\hline
				Method	  &NeRF-SR &NeRF-SR + NeRFLiX++$_{4K-L1}$  \\ \hline
				PSNR$\uparrow$ (dB)    &27.21 &\textbf{29.19}\\
				SSIM$\uparrow$   &0.852 &\textbf{0.908} \\
				LPIPS$\downarrow$   &0.09  &\textbf{0.06} \\ \hline
			\end{tabular}
			\caption{Quantitative improvements over NeRF-SR.}
			\vspace{-0.05in}
			\label{subtab:nerf-sr}
		\end{subtable}
		\caption{Quantitative evaluation of NeRFLiX++$_{4K}$ by comparing it with 4K-NeRF~\cite{wang20224k} and NeRF-SR~\cite{wang2022nerf}. To differentiate between models trained with L1 and GAN supervision, we used subscripts $_{L1}$ and $_{GAN}$, respectively.}
		\label{tab:payoffMatrices}
		
	\end{table}
	
	\vspace{0.03in}
	\noindent\textbf{Comparison to 4K-NeRF.} 
	We also compare NeRFLiX++$_{4K}$ with 4K-NeRF~\cite{wang20224k}\footnote{For a fair comparison, we report the enhanced results using DVGO~\cite{Sun_2022_CVPR} as our baseline, as 4K-NeRF also employs DVGO as its baseline.}. The results in Table~\ref{subtab:4k-nerf} show that NeRFLiX++$_{4K-L1}$ achieves a significant improvement of 0.77dB in PSNR over 4K-NeRF$_{L1}$. Furthermore, NeRFLiX++$_{4K-GAN}$ surpasses 4K-NeRF in terms of perceptual quality. Fig.~\ref{fig:nerf-4k} visually demonstrates that 4K-NeRF$_{GAN}$ fails to reconstruct subtle image structures, while NeRFLiX++$_{4K-GAN}$ effectively restores natural image contents from noisy 1K photos, resulting in superior visual enhancement results.
	
	\vspace{0.03in}
	\noindent\textbf{Comparison to NeRF-SR.} 
	We also compare NeRFLiX++$_{4K-L1}$ with NeRF-SR~\cite{wang2022nerf}. NeRF-SR is a two-stage novel view synthesis approach. In the first stage, they propose a super-sampling NeRF model to generate super-resolved novel views from low-resolution training photos. Then, they utilize a refinement module to enhance the first-stage results. To ensure a fair comparison, we utilize our NeRFLiX++$_{4K-L1}$ model to enhance their first-stage results and quantitatively compare them with their refined results. Table~\ref{subtab:nerf-sr} suggests that NeRFLiX++$_{4K-L1}$ significantly outperforms NeRF-SR, highlighting the effectiveness of our method.
	
	In addition to its superior performance, unlike 4K-NeRF and NeRF-SR models that require re-training for new scenes, NeRFLiX++$_{4K}$ offers the advantage of being NeRF-agnostic and scene-agnostic. This characteristic allows for quick and efficient deployment of NeRFLiX++${4K}$ in various scenarios.
	
	\vspace{0.03in}
	\noindent\textbf{Comparison to existing image and video restorers.}
	Additionally, we compare our NeRFLiX++$_{4K}$ model with state-of-the-art image and video super-resolution approaches, such as SwinIR~\cite{liang2021swinir}, RealESRGAN~\cite{wang2021real}, and RealBasicVSR~\cite{chan2022investigating}. Using TensoRF~\cite{tensorf} as the baseline, we utilize these models to generate enhanced high-resolution images and present the detailed results in Table~\ref{table:4k_sr}. Although these models produce promising restoration outcomes for generally real-world images, they all exhibit inferior performance than our NeRFLiX++$_{4K}$ model, which manifests NeRFLiX++'s excellent restoration capability for NeRF-rendered photos. 
	\begin{table}[t]
		\small
		\setlength{\tabcolsep}{1.6pt}
		\begin{center}
			\begin{tabular}{c|c|c|c|c } 
				
				\hline
				Method                      &SwinIR  &RealESRGAN &RealBasicVSR   &NeRFLiX++$_{4K}$ \\ \hline
				PSNR$\uparrow$ (dB) &23.91 &24.22 &23.97 &\textbf{25.45}\\
				SSIM$\uparrow$      &0.847 &0.860 &0.851 &\textbf{0.885}\\
				LPIPS$\downarrow$     &0.300 &0.299 &0.311 &\textbf{0.184}\\
				\hline 	
			\end{tabular}
		\end{center}
		\vspace{-0.1in}
		\caption{To ensure a fair and objective evaluation, we compare NeRFLiX++$_{4K}$ against various representative general image and video restoration models that are trained using adversarial loss. We assess their abilities of up-sampling and enhancing the rendered views (at a low resolution) of TensoRF~\cite{tensorf}.}
		\vspace{-0.15in}
		\label{table:4k_sr}
	\end{table}

	\vspace{0.03in}
	\noindent\textbf{4K video demo.} We have prepared a video demonstration showcasing the enhancement capabilities of our proposed NeRFLiX++ method, which can be viewed at \url{https://www.youtube.com/watch?v=YiXvgQXiWII}\footnote{We recommend watching it in 4K resolution for the best view.}. It comprises three parts. And the first two segments of the video highlight that while TensoRF and Plenoxels struggle to generate satisfactory 1K novel views, NeRFLiX++ is capable of restoring ultra-high-resolution outputs from these low-resolution noisy views. Notably, NeRFLiX++ even recovers recognizable characters and sharper textures at 4K resolutions. The last segment shows NeRFLiX++ can be used to significantly improve the visual quality of various NeRF models~(\ie ~TensoRF~\cite{tensorf}, Plenoxels~\cite{fridovich2022plenoxels}, RegNeRF~\cite{Niemeyer2021Regnerf}, NLF~\cite{attal2022learning}, DIVeR~\cite{wu2021diver}, NeRF-mm~\cite{wang2021nerf}, etc.).

	\section{Conclusion}
	We introduce NeRFLiX, a general NeRF-agnostic paradigm for high-quality restoration of neural view synthesis. We systematically analyze the NeRF rendering pipeline and introduce the concept of NeRF-style degradations. Towards eliminating NeRF-style artifacts, we present a novel NeRF-style degradation simulator and construct a large-scale simulated dataset. Through training state-of-the-art deep neural networks on the simulated dataset, we successfully remove NeRF artifacts. Additionally, we propose an inter-viewpoint mixer to restore missing details in NeRF-rendered frames by aggregating multi-view frames. Extensive experiments validate the effectiveness of NeRFLiX.
	
	To further enhance the restoration capability and inference efficiency of NeRFLiX, we present NeRFLiX++. It improves upon NeRFLiX by incorporating better degradation modeling and faster inter-viewpoint aggregation techniques. NeRFLiX++ enables realistic 4K view synthesis ability and achieves superior quantitative and qualitative performance, as demonstrated in our extensive experiments.
	
	\vspace{0.1in}
	\noindent\textbf{Acknowledgments.} This work is partially supported by Shenzhen Science and Technology Program KQTD20210811090149095 and also the Pearl River Talent Recruitment Program 2019QN01X226. The work was supported in part by NSFC with Grant No. 62293482, the Basic Research Project No. HZQB-KCZYZ-2021067 of Hetao ShenzhenHK S\&T Cooperation Zone, the National Key R\&D Program of China with grant No. 2018YFB1800800, by Shenzhen Outstanding Talents Training Fund 202002, by Guangdong Research Projects No. 2017ZT07X152 and No. 2019CX01X104, by the Guangdong Provincial Key Laboratory of Future Networks of Intelligence (Grant No. 2022B1212010001), and by Shenzhen Key Laboratory of Big Data and Artificial Intelligence (Grant No. ZDSYS201707251409055). It was also partially supported by NSFC62172348, Outstanding Young Fund of Guangdong Province with No. 2023B1515020055 and Shenzhen General Project with No. JCYJ20220530143604010.

	\appendix 
	
	\begin{figure*}
		\begin{minipage}{0.5\textwidth}
			\centering
			\small
			
			\begin{tabular}{l|c } 
				\hline
				\quad \quad \quad \quad 56Leonard                      & PSNR (dB)${\color{red}\uparrow}$/SSIM${\color{red}\uparrow}$/LPIPS${\color{red}\downarrow}$  \\ \hline
				
				\cellcolor{blue!15}BungeeNeRF ($\frac{1}{2}$) (ECCV'2022)             			 & \cellcolor{blue!15} 21.15 / 0.616 / 0.557  \\ 
				
				\cellcolor{green!15}BungeeNeRF ($\frac{1}{2}$)+ NeRFLiX++$_{4K-L1}$                 & \cellcolor{green!15} {\color{red}22.50} / {\color{red}0.766} / {\color{blue}0.335} \\ 
				\cellcolor{yellow!15}BungeeNeRF ($\frac{1}{2}$)+ NeRFLiX++$_{4K-GAN}$                 & \cellcolor{yellow!15} {\color{blue}21.64} / {\color{blue}0.759} / {\color{red}0.167} \\ \hline  \hline

				\cellcolor{blue!15}BungeeNeRF ($\frac{1}{4}$) (ECCV'2022)             			 &\cellcolor{blue!15} 20.64 / 0.581 / 0.651  \\ 
				
				\cellcolor{green!15}BungeeNeRF ($\frac{1}{4}$)+ NeRFLiX++$_{4K-L1}$                 & \cellcolor{green!15} {\color{red}21.71} / {\color{blue}0.680} / {\color{blue}0.445} \\ 
				\cellcolor{yellow!15}BungeeNeRF ($\frac{1}{4}$)+ NeRFLiX++$_{4K-GAN}$                 &\cellcolor{yellow!15} {\color{blue}21.10} / {\color{red}0.695} / {\color{red}0.268} \\ \hline 
				
				\hline
				\quad \quad \quad \quad Transamerica                      & PSNR (dB)${\color{red}\uparrow}$/SSIM${\color{red}\uparrow}$/LPIPS${\color{red}\downarrow}$  \\ \hline
				
				\cellcolor{blue!15}BungeeNeRF ($\frac{1}{2}$) (ECCV'2022)             			 & \cellcolor{blue!15} 21.50 / 0.585 / 0.562  \\ 
				
				\cellcolor{green!15}BungeeNeRF ($\frac{1}{2}$)+ NeRFLiX++$_{4K-L1}$                 &\cellcolor{green!15} {\color{red}22.83} / {\color{red}0.741} / {\color{blue}0.356}\\ 
				\cellcolor{yellow!15}BungeeNeRF ($\frac{1}{2}$)+ NeRFLiX++$_{4K-GAN}$                 &\cellcolor{yellow!15} {\color{blue}22.04} / {\color{blue}0.738} / {\color{red}0.174} \\ \hline  \hline

				\cellcolor{blue!15}BungeeNeRF ($\frac{1}{4}$) (ECCV'2022)             			 & \cellcolor{blue!15} 21.04 / 0.548 / 0.643  \\ 
				
				\cellcolor{green!15}BungeeNeRF ($\frac{1}{4}$)+ NeRFLiX++$_{4K-L1}$                 & \cellcolor{green!15} {\color{red}22.01} / {\color{blue}0.646} / {\color{blue}0.456} \\ 
				\cellcolor{yellow!15}BungeeNeRF ($\frac{1}{4}$)+ NeRFLiX++$_{4K-GAN}$                 &\cellcolor{yellow!15} {\color{blue}21.45} / {\color{red}0.669} / {\color{red}0.285} \\ \hline 
			\end{tabular}
		\end{minipage}%
		\hfill%
		\begin{minipage}{0.45\textwidth}
			\includegraphics[width=\textwidth]{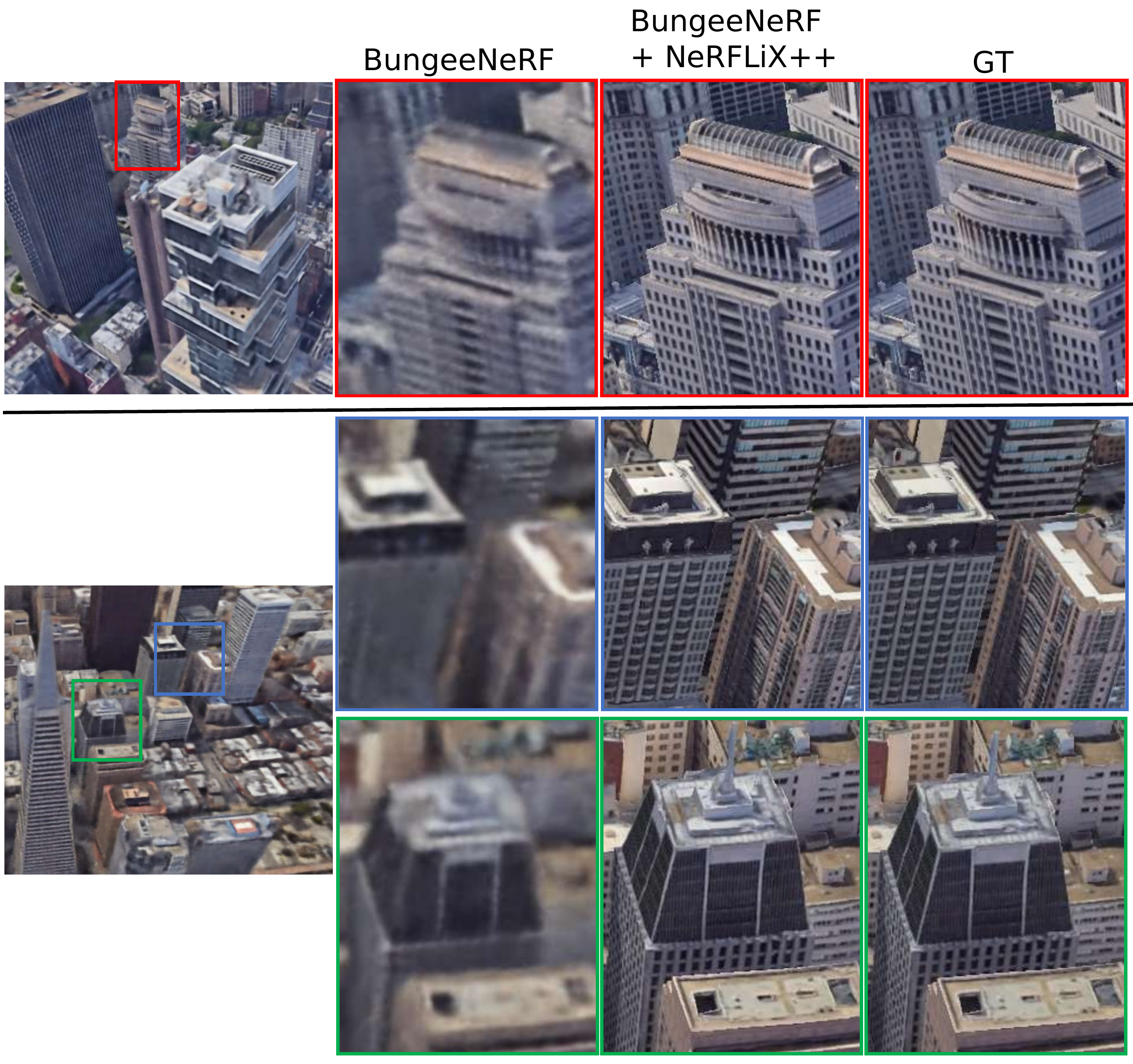}
		\end{minipage}%
		
		\caption{Quantitative and qualitative analysis of adopting NeRFLiX++ for city-scale scenes. We use BungeeNeRF~\cite{xiangli2022bungeenerf} to generate novel views at down-sampling factors of $\{\frac{1}{2},\frac{1}{4}\}$ of the full resolution~($1920\times1080$). The visual examples presented in the right figure demonstrate the strong generalization capability of our proposed NeRFLiX++ on new scenes.}
		\label{fig:bungee}
	\end{figure*}
	\vspace{0.03in}
	\noindent{\color{black}
		{\textbf{Improvements over city-scale BungeeNeRF~\cite{xiangli2022bungeenerf}.} We investigate the potential advantages of utilizing our proposed NeRFLiX++ for city-scale novel view rendering. In our experiments, we employ the official BungeeNeRF model to synthesize two city-scale scenes. As shown in Fig.~\ref{fig:bungee}, it is evident that NeRFLiX++ significantly enhances the quality of city-scale scenes without requiring fine-tuning. This result further underscores the excellent restoration and generalization capabilities of NeRFLiX++. }
	}
	
	\section{NeRF Degradation Simulator}
	\vspace{0.05in}
	\noindent\textbf{Raw data collection.}
	We collect raw sequences from Vimeo90K~\cite{xue2019video} and LLFF-T~\cite{mildenhall2019local}. In total, Vimeo90K contains 64612 7-frame training clips with a $448\times256$ resolution. Three frames~(two reference views and one target view) are selected from a raw sequence of Vimeo90K in a random order. Apart from the inherent displacements within the selected views, we add random global offsets to the two reference views, largely enriching the variety of inter-viewpoint changes.
	On the other hand, we also use the training split of the LLFF dataset, which consists of 8 different forward-facing scenes with 20-62 high-quality input views. Following previous work, we drop the eighth view and use it for evaluation. To construct a training pair from LLFF-T, we randomly select a frame as the target view and then use the proposed view selection algorithm~(Sec.~{\color{red}4.3}) to choose two reference views that are most overlapped with the target view. 
	\begin{figure}[t]
		\centering
		\includegraphics[width=0.8\columnwidth]{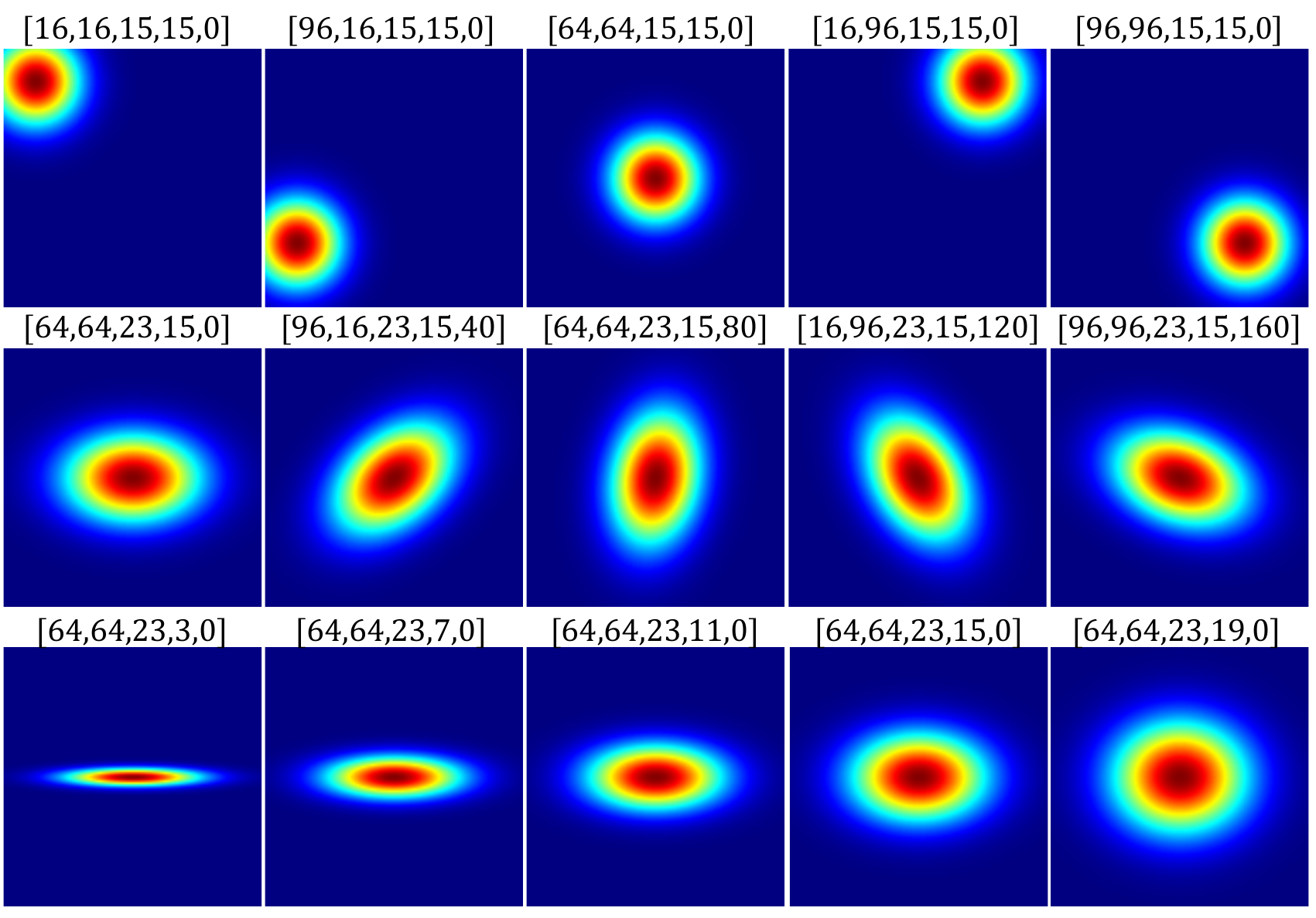} 
		
		\caption{We give some visualized region-adaptive masks. The parameters refer to the values of $[c_i, c_j; \sigma_i, \sigma_j, A]$ in Eq.~({\color{red}3}).
		} 
		\label{fig:ra}
		\vspace{-0.15in}
	\end{figure} %
	
	\vspace{0.05in}
	\noindent\textbf{Hyper-parameter setup.}
	In Eq.~({\color{red}1}), the 2D Gaussian noise map $n$ is generated with a zero mean and a standard deviation ranging from 0.01 to 0.05. The isotropic blur kernel $g$ has a size of $5 \times 5$. We employ a Gaussian blur kernel to produce blurry contents by randomly selecting kernel sizes~(3-7), angles~(0-180), and standard deviations~(0.2-1.2). Last, in order to obtain a region-adaptive blending map $M$ in Eq.~({\color{red}3}), we use random means~($c_i,c_j \in(-16,144)$), standard deviations~($\sigma_i \in (13,25),\sigma_j \in(0,24)$), and orientation angles~($A \in(0,180)$). Additionally, we visualize some generated masks using different hyper-parameter combinations~($[c_i, c_j; \sigma_i, \sigma_j, A]$) in Fig.~\ref{fig:ra}.
	
	\vspace{0.05in}
	\noindent\textbf{Training data size.}
	We investigate the influence of training data size. Under the same training and testing setups, we train several models using different training data sizes. As illustrated in Table~\ref{table:DATA}, we can observe that the final performance is positively correlated with the number of training pairs. Also, we notice the IVM trained with only LLFF-T data or additional few simulated pairs~(10$\%$ of the Vimeo90K) fails to enhance the TensoRF-rendered results, \ie, there is no obvious improvement compared to TensoRF~\cite{tensorf}. This experiment demonstrates the importance of sizable training pairs for training a NeRF restorer.
	
	\begin{table}[t]
		\small
		\setlength{\tabcolsep}{5pt}
		\begin{center}
			
			
			\begin{tabular}{l|c|c|c|c|c } 
				\hline
				Settings  & 10$\%$ & 50$\%$ & 100$\%$ &PSNR~(dB)&SSIM\\ \hline
				LLFF-T       & & & &26.28& 0.837 \\
				LLFF-T+       &\cmark & & &26.71& 0.840 \\
				LLFF-T+       & &\cmark & &27.08& 0.856 \\
				LLFF-T+       & & &\cmark &\textbf{27.39}&\textbf{ 0.867} \\ \hline \hline 
				TensoRF (Base)  &- &- &- &26.70 &0.838 \\ \hline

			\end{tabular}
		\end{center}
		\vspace{-0.2in}
		\caption{Quantitative results of different training data sizes. First, we train an IVM model only using the LLFF-T. Then, we gradually increase the simulated pairs~(10$\%$,  50$\%$, 100$\%$) from Vimeo90K~\cite{xue2019video} to train another three IVM models. }
		\vspace{-0.15in}
		\label{table:DATA}
		
	\end{table}
	\begin{figure*}[t]
		\begin{center}
			\includegraphics[width=\linewidth]{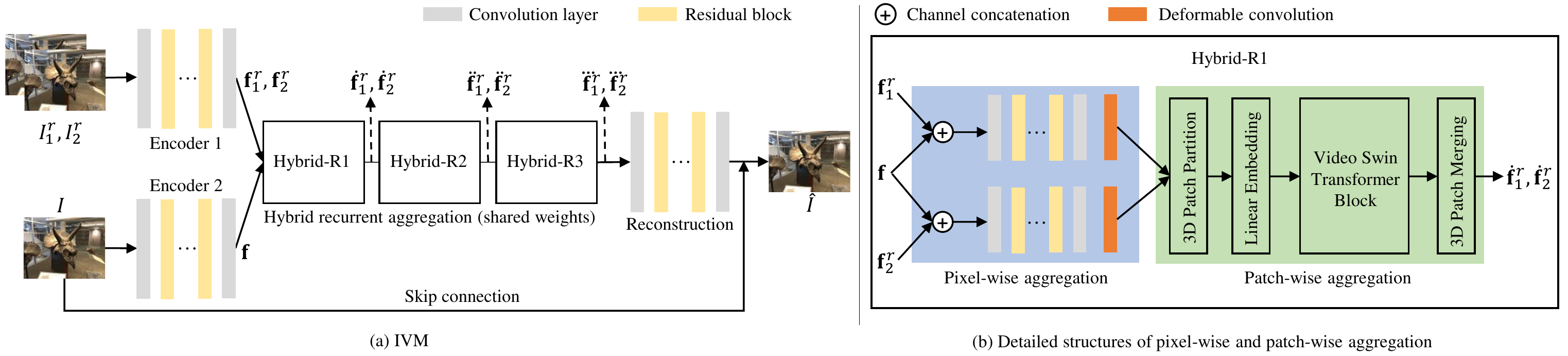} 
		\end{center}
		\vspace{-0.15in}
		\caption{The detailed framework architecture of our proposed IVM.
			\vspace{-0.1in}
		}
		\label{fig:framework_ivm}
	\end{figure*}
	\vspace{-0.05in}
	\section{Inter-viewpoint Mixer}
	\vspace{0.05in}
	In Sec.~{\color{red}4.2}, we briefly describe the framework architecture of our inter-viewpoint mixer (IVM). Here we provide more details. As illustrated in Fig.~\ref{fig:framework_ivm}{\color{red}a}, there are two convolutional modules~(``Encoder 1/2") to extract features of the degraded view $I$ and its two reference views $\{I_1^r,I_2^r\}$, respectively. Then, we develop a hybrid recurrent aggregation module that iteratively performs pixel-wise and patch-wise fusion. At last, a reconstruction module is implemented by a sequence of residual blocks~(40 blocks) to output the enhanced view ${\hat I}$. The default channel size is 128.
	
	\vspace{0.05in}
	\noindent\textbf{Feature extraction.}
	Given a rendered view $I$ and its two reference views ${I}_{1,2}^r$, we aim to utilize the two encoders to extract deep image features $\mathbf{f}$ and $\mathbf{f}_{1,2}^r$, respectively. As detailed in Fig.~\ref{fig:framework_ivm}{\color{red}a}, the two encoders share an identical structure. A convolutional layer is first adopted to convert an RGB frame to a high-dimensional feature. Then we further extract the deep image feature using 5 stacked residual blocks followed by another convolution layer. 
	
	\vspace{0.05in}
	\noindent\textbf{Hybrid recurrent aggregation.} As depicted in Fig.~\ref{fig:framework_ivm}{\color{red}a}, we employ three hybrid recurrent aggregation blocks~(termed ``Hybrid-R1(2,3)") to progressively fuse the inter-viewpoint information from the image features~($\mathbf{f}$ and $\mathbf{f}_{\{1,2\}}^r$). Next, we take the first iteration as an example to illustrate our aggregation scheme.
	
	\vspace{0.05in}
	\noindent\textbf{Pixel-wise aggregation.}
	As shown in Fig.~\ref{fig:framework_ivm}{\color{red}b}, we first merge the target view feature $\mathbf{f}$ and one of the reference features~$\mathbf{f}_{\{1,2\}}^r$ by channel concatenation. Then we use a convolutional layer to reduce the channel dimension and five residual blocks followed by another convolutional layer to obtain a fused deep feature. Later on, the fused feature and the reference feature are further aggregated through a deformable convolution. And the other reference image follows the same processing pipeline. In this case, we finally obtain two features after the pixel-wise aggregation.
	
	\vspace{0.05in}
	\noindent\textbf{Patch-wise aggregation.}
	We adopt a window-based attention mechanism~\cite{liu2022video} to accomplish patch-wise aggregation. In detail, the pixel-wisely fused features are first divided into several 3D slices through a 3D patch partition layer. Then, we obtain 3D tokens via a linear embedding operation and aggregate patch-wise information using a video Swin transformer block. Finally, 3D patches are regrouped into a 3D feature map.
	
	In the next iteration, we split the 3D feature map into two ``reference" features $\mathbf{\dot{f}}_{\{1,2\}}^r$ and repeat the pixel-wise and patch-wise aggregation. Note that, the weights of pixel-wise and patch-wise modules are shared across all iterations to reduce the model complexity.

	{\small
		\bibliographystyle{ieee_fullname}
		\bibliography{NeRFLiX_Plus_finalversion}

\begin{thebibliography}{10}\itemsep=-1pt

\bibitem{attal2022learning}
Benjamin Attal, Jia-Bin Huang, Michael Zollh{\"o}fer, Johannes Kopf, and
  Changil Kim.
\newblock Learning neural light fields with ray-space embedding networks.
\newblock In {\em Proceedings of the IEEE/CVF Conference on Computer Vision and
  Pattern Recognition (CVPR)}, 2022.

\bibitem{barron2021mip}
Jonathan~T Barron, Ben Mildenhall, Matthew Tancik, Peter Hedman, Ricardo
  Martin-Brualla, and Pratul~P Srinivasan.
\newblock Mip-nerf: A multiscale representation for anti-aliasing neural
  radiance fields.
\newblock In {\em Proceedings of the IEEE/CVF International Conference on
  Computer Vision}, pages 5855--5864, 2021.

\bibitem{cao2021vsrt}
Jiezhang Cao, Yawei Li, Kai Zhang, and Luc Van~Gool.
\newblock Video super-resolution transformer.
\newblock {\em arXiv}, 2021.

\bibitem{cao2022datsr}
Jiezhang Cao, Jingyun Liang, Kai Zhang, Yawei Li, Yulun Zhang, Wenguan Wang,
  and Luc Van~Gool.
\newblock Reference-based image super-resolution with deformable attention
  transformer.
\newblock In {\em European conference on computer vision}, 2022.

\bibitem{cao2022vdtr}
Mingden Cao, Yanbo Fan, Yong Zhang, Jue Wang, and Yujiu Yang.
\newblock Vdtr: Video deblurring with transformer.
\newblock {\em IEEE Transactions on Circuits and Systems for Video Technology},
  2022.

\bibitem{chan2021basicvsr}
Kelvin~CK Chan, Xintao Wang, Ke Yu, Chao Dong, and Chen~Change Loy.
\newblock Basicvsr: The search for essential components in video
  super-resolution and beyond.
\newblock In {\em Proceedings of the IEEE/CVF Conference on Computer Vision and
  Pattern Recognition}, 2021.

\bibitem{chan2021basicvsr++}
Kelvin~CK Chan, Shangchen Zhou, Xiangyu Xu, and Chen~Change Loy.
\newblock Basicvsr++: Improving video super-resolution with enhanced
  propagation and alignment.
\newblock In {\em Proceedings of the IEEE/CVF Conference on Computer Vision and
  Pattern Recognition}, pages 5972--5981, 2022.

\bibitem{chan2022investigating}
Kelvin~CK Chan, Shangchen Zhou, Xiangyu Xu, and Chen~Change Loy.
\newblock Investigating tradeoffs in real-world video super-resolution.
\newblock In {\em Proceedings of the IEEE/CVF Conference on Computer Vision and
  Pattern Recognition}, pages 5962--5971, 2022.

\bibitem{tensorf}
Anpei Chen, Zexiang Xu, Andreas Geiger, Jingyi Yu, and Hao Su.
\newblock Tensorf: Tensorial radiance fields.
\newblock In {\em European Conference on Computer Vision (ECCV)}, 2022.

\bibitem{chen2021mvsnerf}
Anpei Chen, Zexiang Xu, Fuqiang Zhao, Xiaoshuai Zhang, Fanbo Xiang, Jingyi Yu,
  and Hao Su.
\newblock Mvsnerf: Fast generalizable radiance field reconstruction from
  multi-view stereo.
\newblock In {\em Proceedings of the IEEE/CVF International Conference on
  Computer Vision}, pages 14124--14133, 2021.

\bibitem{chen2022mobilenerf}
Zhiqin Chen, Thomas Funkhouser, Peter Hedman, and Andrea Tagliasacchi.
\newblock Mobilenerf: Exploiting the polygon rasterization pipeline for
  efficient neural field rendering on mobile architectures.
\newblock {\em arXiv preprint arXiv:2208.00277}, 2022.

\bibitem{cole2021differentiable}
Forrester Cole, Kyle Genova, Avneesh Sud, Daniel Vlasic, and Zhoutong Zhang.
\newblock Differentiable surface rendering via non-differentiable sampling.
\newblock In {\em Proceedings of the IEEE/CVF International Conference on
  Computer Vision}, pages 6088--6097, 2021.

\bibitem{dai2017deformable}
Jifeng Dai, Haozhi Qi, Yuwen Xiong, Yi Li, Guodong Zhang, Han Hu, and Yichen
  Wei.
\newblock Deformable convolutional networks.
\newblock In {\em Proceedings of the IEEE International Conference on Computer
  Vision}, pages 764--773, 2017.

\bibitem{deng2022compressing}
Chenxi~Lola Deng and Enzo Tartaglione.
\newblock Compressing explicit voxel grid representations: fast nerfs become
  also small.
\newblock {\em arXiv preprint arXiv:2210.12782}, 2022.

\bibitem{deng2022depth}
Kangle Deng, Andrew Liu, Jun-Yan Zhu, and Deva Ramanan.
\newblock Depth-supervised nerf: Fewer views and faster training for free.
\newblock In {\em Proceedings of the IEEE/CVF Conference on Computer Vision and
  Pattern Recognition}, pages 12882--12891, 2022.

\bibitem{dong2015image}
Chao Dong, Chen~Change Loy, Kaiming He, and Xiaoou Tang.
\newblock Image super-resolution using deep convolutional networks.
\newblock {\em IEEE transactions on pattern analysis and machine intelligence},
  38(2):295--307, 2015.

\bibitem{fridovich2022plenoxels}
Sara Fridovich-Keil, Alex Yu, Matthew Tancik, Qinhong Chen, Benjamin Recht, and
  Angjoo Kanazawa.
\newblock Plenoxels: Radiance fields without neural networks.
\newblock In {\em Proceedings of the IEEE/CVF Conference on Computer Vision and
  Pattern Recognition}, pages 5501--5510, 2022.

\bibitem{garbin2021fastnerf}
Stephan~J Garbin, Marek Kowalski, Matthew Johnson, Jamie Shotton, and Julien
  Valentin.
\newblock Fastnerf: High-fidelity neural rendering at 200fps.
\newblock In {\em Proceedings of the IEEE/CVF International Conference on
  Computer Vision}, pages 14346--14355, 2021.

\bibitem{geusebroek2003fast}
J-M Geusebroek, Arnold~WM Smeulders, and Joost Van De~Weijer.
\newblock Fast anisotropic gauss filtering.
\newblock {\em IEEE transactions on image processing}, 12(8):938--943, 2003.

\bibitem{guo2022nerfren}
Yuan-Chen Guo, Di Kang, Linchao Bao, Yu He, and Song-Hai Zhang.
\newblock Nerfren: Neural radiance fields with reflections.
\newblock In {\em Proceedings of the IEEE/CVF Conference on Computer Vision and
  Pattern Recognition}, pages 18409--18418, 2022.

\bibitem{hu2022efficientnerf}
Tao Hu, Shu Liu, Yilun Chen, Tiancheng Shen, and Jiaya Jia.
\newblock Efficientnerf: Efficient neural radiance fields.
\newblock In {\em Proceedings of the IEEE/CVF Conference on Computer Vision and
  Pattern Recognition}, pages 12902--12911, 2022.

\bibitem{huang2023refsr}
Xudong Huang, Wei Li, Jie Hu, Hanting Chen, and Yunhe Wang.
\newblock Refsr-nerf: Towards high fidelity and super resolution view
  synthesis.
\newblock In {\em Proceedings of the IEEE/CVF Conference on Computer Vision and
  Pattern Recognition}, pages 8244--8253, 2023.

\bibitem{ichnowski2021dex}
Jeffrey Ichnowski*, Yahav Avigal*, Justin Kerr, and Ken Goldberg.
\newblock {Dex-NeRF}: Using a neural radiance field to grasp transparent
  objects.
\newblock In {\em Conference on Robot Learning (CoRL)}, 2020.

\bibitem{jeong2021self}
Yoonwoo Jeong, Seokjun Ahn, Christopher Choy, Anima Anandkumar, Minsu Cho, and
  Jaesik Park.
\newblock Self-calibrating neural radiance fields.
\newblock In {\em Proceedings of the IEEE/CVF International Conference on
  Computer Vision}, pages 5846--5854, 2021.

\bibitem{johari2022geonerf}
Mohammad~Mahdi Johari, Yann Lepoittevin, and Fran{\c{c}}ois Fleuret.
\newblock Geonerf: Generalizing nerf with geometry priors.
\newblock In {\em Proceedings of the IEEE/CVF Conference on Computer Vision and
  Pattern Recognition}, pages 18365--18375, 2022.

\bibitem{knapitsch2017tanks}
Arno Knapitsch, Jaesik Park, Qian-Yi Zhou, and Vladlen Koltun.
\newblock Tanks and temples: Benchmarking large-scale scene reconstruction.
\newblock {\em ACM Transactions on Graphics (ToG)}, 36(4):1--13, 2017.

\bibitem{kurz2022adanerf}
Andreas Kurz, Thomas Neff, Zhaoyang Lv, Michael Zollh{\"o}fer, and Markus
  Steinberger.
\newblock Adanerf: Adaptive sampling for real-time rendering of neural radiance
  fields.
\newblock In {\em European Conference on Computer Vision}, pages 254--270.
  Springer, 2022.

\bibitem{li2020mucan}
Wenbo Li, Xin Tao, Taian Guo, Lu Qi, Jiangbo Lu, and Jiaya Jia.
\newblock Mucan: Multi-correspondence aggregation network for video
  super-resolution.
\newblock In {\em ECCV}, pages 335--351. Springer, 2020.

\bibitem{li2020lapar}
Wenbo Li, Kun Zhou, Lu Qi, Nianjuan Jiang, Jiangbo Lu, and Jiaya Jia.
\newblock Lapar: Linearly-assembled pixel-adaptive regression network for
  single image super-resolution and beyond.
\newblock {\em Advances in Neural Information Processing Systems},
  33:20343--20355, 2020.

\bibitem{li2022best}
Wenbo Li, Kun Zhou, Lu Qi, Liying Lu, and Jiangbo Lu.
\newblock Best-buddy gans for highly detailed image super-resolution.
\newblock In {\em Proceedings of the AAAI Conference on Artificial
  Intelligence}, volume~36, pages 1412--1420, 2022.

\bibitem{li2019feedback}
Zhen Li, Jinglei Yang, Zheng Liu, Xiaomin Yang, Gwanggil Jeon, and Wei Wu.
\newblock Feedback network for image super-resolution.
\newblock In {\em Proceedings of the IEEE/CVF conference on computer vision and
  pattern recognition}, pages 3867--3876, 2019.

\bibitem{liang2022vrt}
Jingyun Liang, Jiezhang Cao, Yuchen Fan, Kai Zhang, Rakesh Ranjan, Yawei Li,
  Radu Timofte, and Luc Van~Gool.
\newblock Vrt: A video restoration transformer.
\newblock {\em arXiv preprint arXiv:2201.12288}, 2022.

\bibitem{liang2021swinir}
Jingyun Liang, Jiezhang Cao, Guolei Sun, Kai Zhang, Luc Van~Gool, and Radu
  Timofte.
\newblock Swinir: Image restoration using swin transformer.
\newblock In {\em Proceedings of the IEEE/CVF International Conference on
  Computer Vision}, pages 1833--1844, 2021.

\bibitem{lin2021barf}
Chen-Hsuan Lin, Wei-Chiu Ma, Antonio Torralba, and Simon Lucey.
\newblock Barf: Bundle-adjusting neural radiance fields.
\newblock In {\em Proceedings of the IEEE/CVF International Conference on
  Computer Vision}, pages 5741--5751, 2021.

\bibitem{lin2021efficient}
Haotong Lin, Sida Peng, Zhen Xu, Hujun Bao, and Xiaowei Zhou.
\newblock Efficient neural radiance fields with learned depth-guided sampling.
\newblock {\em arXiv preprint arXiv:2112.01517}, 2021.

\bibitem{liu2022video}
Ze Liu, Jia Ning, Yue Cao, Yixuan Wei, Zheng Zhang, Stephen Lin, and Han Hu.
\newblock Video swin transformer.
\newblock In {\em Proceedings of the IEEE/CVF Conference on Computer Vision and
  Pattern Recognition}, pages 3202--3211, 2022.

\bibitem{martin2021nerf}
Ricardo Martin-Brualla, Noha Radwan, Mehdi~SM Sajjadi, Jonathan~T Barron,
  Alexey Dosovitskiy, and Daniel Duckworth.
\newblock Nerf in the wild: Neural radiance fields for unconstrained photo
  collections.
\newblock In {\em Proceedings of the IEEE/CVF Conference on Computer Vision and
  Pattern Recognition}, pages 7210--7219, 2021.

\bibitem{mildenhall2022nerf}
Ben Mildenhall, Peter Hedman, Ricardo Martin-Brualla, Pratul~P Srinivasan, and
  Jonathan~T Barron.
\newblock Nerf in the dark: High dynamic range view synthesis from noisy raw
  images.
\newblock In {\em Proceedings of the IEEE/CVF Conference on Computer Vision and
  Pattern Recognition}, pages 16190--16199, 2022.

\bibitem{mildenhall2019local}
Ben Mildenhall, Pratul~P Srinivasan, Rodrigo Ortiz-Cayon, Nima~Khademi
  Kalantari, Ravi Ramamoorthi, Ren Ng, and Abhishek Kar.
\newblock Local light field fusion: Practical view synthesis with prescriptive
  sampling guidelines.
\newblock {\em ACM Transactions on Graphics (TOG)}, 38(4):1--14, 2019.

\bibitem{mildenhall2020nerf}
Ben Mildenhall, Pratul~P. Srinivasan, Matthew Tancik, Jonathan~T. Barron, Ravi
  Ramamoorthi, and Ren Ng.
\newblock Nerf: Representing scenes as neural radiance fields for view
  synthesis.
\newblock In {\em ECCV}, 2020.

\bibitem{mueller2022instant}
Thomas M\"uller, Alex Evans, Christoph Schied, and Alexander Keller.
\newblock Instant neural graphics primitives with a multiresolution hash
  encoding.
\newblock {\em ACM Trans. Graph.}, 41(4):102:1--102:15, July 2022.

\bibitem{Niemeyer2021Regnerf}
Michael Niemeyer, Jonathan~T. Barron, Ben Mildenhall, Mehdi S.~M. Sajjadi,
  Andreas Geiger, and Noha Radwan.
\newblock Regnerf: Regularizing neural radiance fields for view synthesis from
  sparse inputs.
\newblock In {\em Proc. IEEE Conf. on Computer Vision and Pattern Recognition
  (CVPR)}, 2022.

\bibitem{pumarola2021d}
Albert Pumarola, Enric Corona, Gerard Pons-Moll, and Francesc Moreno-Noguer.
\newblock D-nerf: Neural radiance fields for dynamic scenes.
\newblock In {\em Proceedings of the IEEE/CVF Conference on Computer Vision and
  Pattern Recognition}, pages 10318--10327, 2021.

\bibitem{ranjan2017optical}
Anurag Ranjan and Michael~J Black.
\newblock Optical flow estimation using a spatial pyramid network.
\newblock In {\em Proceedings of the IEEE/CVF Conference on Computer Vision and
  Pattern Recognition}, pages 4161--4170, 2017.

\bibitem{rebain2021derf}
Daniel Rebain, Wei Jiang, Soroosh Yazdani, Ke Li, Kwang~Moo Yi, and Andrea
  Tagliasacchi.
\newblock Derf: Decomposed radiance fields.
\newblock In {\em Proceedings of the IEEE/CVF Conference on Computer Vision and
  Pattern Recognition}, pages 14153--14161, 2021.

\bibitem{reiser2021kilonerf}
Christian Reiser, Songyou Peng, Yiyi Liao, and Andreas Geiger.
\newblock Kilonerf: Speeding up neural radiance fields with thousands of tiny
  mlps.
\newblock In {\em Proceedings of the IEEE/CVF International Conference on
  Computer Vision}, pages 14335--14345, 2021.

\bibitem{suhail2022light}
Mohammed Suhail, Carlos Esteves, Leonid Sigal, and Ameesh Makadia.
\newblock Light field neural rendering.
\newblock In {\em Proceedings of the IEEE/CVF Conference on Computer Vision and
  Pattern Recognition}, pages 8269--8279, 2022.

\bibitem{Sun_2022_CVPR}
Cheng Sun, Min Sun, and Hwann-Tzong Chen.
\newblock Direct voxel grid optimization: Super-fast convergence for radiance
  fields reconstruction.
\newblock In {\em Proceedings of the IEEE/CVF Conference on Computer Vision and
  Pattern Recognition (CVPR)}, pages 5459--5469, June 2022.

\bibitem{szegedy2016rethinking}
Christian Szegedy, Vincent Vanhoucke, Sergey Ioffe, Jon Shlens, and Zbigniew
  Wojna.
\newblock Rethinking the inception architecture for computer vision.
\newblock In {\em Proceedings of the IEEE/CVF Conference on Computer Vision and
  Pattern Recognition}, pages 2818--2826, 2016.

\bibitem{tancik2022block}
Matthew Tancik, Vincent Casser, Xinchen Yan, Sabeek Pradhan, Ben Mildenhall,
  Pratul~P Srinivasan, Jonathan~T Barron, and Henrik Kretzschmar.
\newblock Block-nerf: Scalable large scene neural view synthesis.
\newblock In {\em Proceedings of the IEEE/CVF Conference on Computer Vision and
  Pattern Recognition}, pages 8248--8258, 2022.

\bibitem{teed2020raft}
Zachary Teed and Jia Deng.
\newblock Raft: Recurrent all-pairs field transforms for optical flow.
\newblock In {\em European conference on computer vision}, pages 402--419.
  Springer, 2020.

\bibitem{tian2020tdan}
Yapeng Tian, Yulun Zhang, Yun Fu, and Chenliang Xu.
\newblock Tdan: Temporally-deformable alignment network for video
  super-resolution.
\newblock In {\em Proceedings of the IEEE/CVF Conference on Computer Vision and
  Pattern Recognition}, pages 3360--3369, 2020.

\bibitem{van2008visualizing}
Laurens Van~der Maaten and Geoffrey Hinton.
\newblock Visualizing data using t-sne.
\newblock {\em Journal of machine learning research}, 9(11), 2008.

\bibitem{wang2022nerf}
Chen Wang, Xian Wu, Yuan-Chen Guo, Song-Hai Zhang, Yu-Wing Tai, and Shi-Min Hu.
\newblock Nerf-sr: High quality neural radiance fields using supersampling.
\newblock In {\em Proceedings of the 30th ACM International Conference on
  Multimedia}, pages 6445--6454, 2022.

\bibitem{wang2018learning}
Longguang Wang, Yulan Guo, Zaiping Lin, Xinpu Deng, and Wei An.
\newblock Learning for video super-resolution through hr optical flow
  estimation.
\newblock In {\em Asian Conference on Computer Vision}, pages 514--529.
  Springer, 2018.

\bibitem{wang2021ibrnet}
Qianqian Wang, Zhicheng Wang, Kyle Genova, Pratul~P Srinivasan, Howard Zhou,
  Jonathan~T Barron, Ricardo Martin-Brualla, Noah Snavely, and Thomas
  Funkhouser.
\newblock Ibrnet: Learning multi-view image-based rendering.
\newblock In {\em Proceedings of the IEEE/CVF Conference on Computer Vision and
  Pattern Recognition}, pages 4690--4699, 2021.

\bibitem{wang2019edvr}
Xintao Wang, Kelvin~CK Chan, Ke Yu, Chao Dong, and Chen Change~Loy.
\newblock Edvr: Video restoration with enhanced deformable convolutional
  networks.
\newblock In {\em Proceedings of the IEEE/CVF Conference on Computer Vision and
  Pattern Recognition Workshops}, pages 0--0, 2019.

\bibitem{wang2021real}
Xintao Wang, Liangbin Xie, Chao Dong, and Ying Shan.
\newblock Real-esrgan: Training real-world blind super-resolution with pure
  synthetic data.
\newblock In {\em Proceedings of the IEEE/CVF International Conference on
  Computer Vision}, pages 1905--1914, 2021.

\bibitem{wang2018esrgan}
Xintao Wang, Ke Yu, Shixiang Wu, Jinjin Gu, Yihao Liu, Chao Dong, Yu Qiao, and
  Chen Change~Loy.
\newblock Esrgan: Enhanced super-resolution generative adversarial networks.
\newblock In {\em Proceedings of the European conference on computer vision
  (ECCV) workshops}, pages 0--0, 2018.

\bibitem{wang2004image}
Zhou Wang, Alan~C Bovik, Hamid~R Sheikh, and Eero~P Simoncelli.
\newblock Image quality assessment: from error visibility to structural
  similarity.
\newblock {\em IEEE transactions on image processing}, 13(4):600--612, 2004.

\bibitem{wang20224k}
Zhongshu Wang, Lingzhi Li, Zhen Shen, Li Shen, and Liefeng Bo.
\newblock 4k-nerf: High fidelity neural radiance fields at ultra high
  resolutions.
\newblock {\em arXiv preprint arXiv:2212.04701}, 2022.

\bibitem{wang2021nerf}
Zirui Wang, Shangzhe Wu, Weidi Xie, Min Chen, and Victor~Adrian Prisacariu.
\newblock Nerf--: Neural radiance fields without known camera parameters.
\newblock {\em arXiv preprint arXiv:2102.07064}, 2021.

\bibitem{wu2021diver}
Liwen Wu, Jae~Yong Lee, Anand Bhattad, Yu-Xiong Wang, and David Forsyth.
\newblock Diver: Real-time and accurate neural radiance fields with
  deterministic integration for volume rendering.
\newblock In {\em Proceedings of the IEEE/CVF Conference on Computer Vision and
  Pattern Recognition}, pages 16200--16209, 2022.

\bibitem{wu2022dof}
Zijin Wu, Xingyi Li, Juewen Peng, Hao Lu, Zhiguo Cao, and Weicai Zhong.
\newblock Dof-nerf: Depth-of-field meets neural radiance fields.
\newblock In {\em Proceedings of the 30th ACM International Conference on
  Multimedia}, pages 1718--1729, 2022.

\bibitem{xiang2021neutex}
Fanbo Xiang, Zexiang Xu, Milos Hasan, Yannick Hold-Geoffroy, Kalyan Sunkavalli,
  and Hao Su.
\newblock Neutex: Neural texture mapping for volumetric neural rendering.
\newblock In {\em Proceedings of the IEEE/CVF Conference on Computer Vision and
  Pattern Recognition}, pages 7119--7128, 2021.

\bibitem{xiangli2022bungeenerf}
Yuanbo Xiangli, Linning Xu, Xingang Pan, Nanxuan Zhao, Anyi Rao, Christian
  Theobalt, Bo Dai, and Dahua Lin.
\newblock Bungeenerf: Progressive neural radiance field for extreme multi-scale
  scene rendering.
\newblock In {\em European conference on computer vision}, pages 106--122.
  Springer, 2022.

\bibitem{xu2022point}
Qiangeng Xu, Zexiang Xu, Julien Philip, Sai Bi, Zhixin Shu, Kalyan Sunkavalli,
  and Ulrich Neumann.
\newblock Point-nerf: Point-based neural radiance fields.
\newblock In {\em Proceedings of the IEEE/CVF Conference on Computer Vision and
  Pattern Recognition}, pages 5438--5448, 2022.

\bibitem{xue2019video}
Tianfan Xue, Baian Chen, Jiajun Wu, Donglai Wei, and William~T Freeman.
\newblock Video enhancement with task-oriented flow.
\newblock {\em IJCV}, 127(8):1106--1125, 2019.

\bibitem{yang2022recursive}
Guo-Wei Yang, Wen-Yang Zhou, Hao-Yang Peng, Dun Liang, Tai-Jiang Mu, and
  Shi-Min Hu.
\newblock Recursive-nerf: An efficient and dynamically growing nerf.
\newblock {\em IEEE Transactions on Visualization and Computer Graphics}, 2022.

\bibitem{yariv2020multiview}
Lior Yariv, Yoni Kasten, Dror Moran, Meirav Galun, Matan Atzmon, Basri Ronen,
  and Yaron Lipman.
\newblock Multiview neural surface reconstruction by disentangling geometry and
  appearance.
\newblock {\em Advances in Neural Information Processing Systems},
  33:2492--2502, 2020.

\bibitem{yoon2023cross}
Youngho Yoon and Kuk-Jin Yoon.
\newblock Cross-guided optimization of radiance fields with multi-view image
  super-resolution for high-resolution novel view synthesis.
\newblock In {\em Proceedings of the IEEE/CVF Conference on Computer Vision and
  Pattern Recognition}, pages 12428--12438, 2023.

\bibitem{yu2021plenoctrees}
Alex Yu, Ruilong Li, Matthew Tancik, Hao Li, Ren Ng, and Angjoo Kanazawa.
\newblock Plenoctrees for real-time rendering of neural radiance fields.
\newblock In {\em Proceedings of the IEEE/CVF International Conference on
  Computer Vision}, pages 5752--5761, 2021.

\bibitem{yu2021path}
Ke Yu, Xintao Wang, Chao Dong, Xiaoou Tang, and Chen~Change Loy.
\newblock Path-restore: Learning network path selection for image restoration.
\newblock {\em IEEE Transactions on Pattern Analysis and Machine Intelligence},
  2021.

\bibitem{yu2020joint}
Songhyun Yu, Bumjun Park, Junwoo Park, and Jechang Jeong.
\newblock Joint learning of blind video denoising and optical flow estimation.
\newblock In {\em Proceedings of the IEEE/CVF Conference on Computer Vision and
  Pattern Recognition Workshops}, pages 500--501, 2020.

\bibitem{zhang2022vmrf}
Jiahui Zhang, Fangneng Zhan, Rongliang Wu, Yingchen Yu, Wenqing Zhang, Bai
  Song, Xiaoqin Zhang, and Shijian Lu.
\newblock Vmrf: View matching neural radiance fields.
\newblock In {\em Proceedings of the 30th ACM International Conference on
  Multimedia}, pages 6579--6587, 2022.

\bibitem{zhang2020deep}
Kai Zhang, Luc~Van Gool, and Radu Timofte.
\newblock Deep unfolding network for image super-resolution.
\newblock In {\em Proceedings of the IEEE/CVF conference on computer vision and
  pattern recognition}, pages 3217--3226, 2020.

\bibitem{zhang2021designing}
Kai Zhang, Jingyun Liang, Luc Van~Gool, and Radu Timofte.
\newblock Designing a practical degradation model for deep blind image
  super-resolution.
\newblock In {\em Proceedings of the IEEE/CVF International Conference on
  Computer Vision}, pages 4791--4800, 2021.

\bibitem{zhang2021physg}
Kai Zhang, Fujun Luan, Qianqian Wang, Kavita Bala, and Noah Snavely.
\newblock Physg: Inverse rendering with spherical gaussians for physics-based
  material editing and relighting.
\newblock In {\em Proceedings of the IEEE/CVF Conference on Computer Vision and
  Pattern Recognition}, pages 5453--5462, 2021.

\bibitem{zhang2020nerf++}
Kai Zhang, Gernot Riegler, Noah Snavely, and Vladlen Koltun.
\newblock Nerf++: Analyzing and improving neural radiance fields.
\newblock {\em arXiv preprint arXiv:2010.07492}, 2020.

\bibitem{zhang2018unreasonable}
Richard Zhang, Phillip Isola, Alexei~A Efros, Eli Shechtman, and Oliver Wang.
\newblock The unreasonable effectiveness of deep features as a perceptual
  metric.
\newblock In {\em Proceedings of the IEEE/CVF Conference on Computer Vision and
  Pattern Recognition}, pages 586--595, 2018.

\bibitem{zhang2022fast}
Wenyuan Zhang, Ruofan Xing, Yunfan Zeng, Yu-Shen Liu, Kanle Shi, and Zhizhong
  Han.
\newblock Fast learning radiance fields by shooting much fewer rays.
\newblock {\em arXiv preprint arXiv:2208.06821}, 2022.

\bibitem{zhang2021nerfactor}
Xiuming Zhang, Pratul~P Srinivasan, Boyang Deng, Paul Debevec, William~T
  Freeman, and Jonathan~T Barron.
\newblock Nerfactor: Neural factorization of shape and reflectance under an
  unknown illumination.
\newblock {\em ACM Transactions on Graphics (TOG)}, 40(6):1--18, 2021.

\bibitem{zhang2022modeling}
Yuanqing Zhang, Jiaming Sun, Xingyi He, Huan Fu, Rongfei Jia, and Xiaowei Zhou.
\newblock Modeling indirect illumination for inverse rendering.
\newblock In {\em Proceedings of the IEEE/CVF Conference on Computer Vision and
  Pattern Recognition}, pages 18643--18652, 2022.

\bibitem{zhang2018residual}
Yulun Zhang, Yapeng Tian, Yu Kong, Bineng Zhong, and Yun Fu.
\newblock Residual dense network for image super-resolution.
\newblock In {\em Proceedings of the IEEE/CVF Conference on Computer Vision and
  Pattern Recognition}, pages 2472--2481, 2018.

\bibitem{zhou2022revisiting}
Kun Zhou, Wenbo Li, Liying Lu, Xiaoguang Han, and Jiangbo Lu.
\newblock Revisiting temporal alignment for video restoration.
\newblock In {\em Proceedings of the IEEE/CVF Conference on Computer Vision and
  Pattern Recognition}, pages 6053--6062, 2022.

\bibitem{Zhou_2023_CVPR}
Kun Zhou, Wenbo Li, Yi Wang, Tao Hu, Nianjuan Jiang, Xiaoguang Han, and Jiangbo
  Lu.
\newblock Nerflix: High-quality neural view synthesis by learning a
  degradation-driven inter-viewpoint mixer.
\newblock In {\em Proceedings of the IEEE/CVF Conference on Computer Vision and
  Pattern Recognition (CVPR)}, pages 12363--12374, June 2023.

\end{thebibliography}
	}

\end{document}